\documentclass{article}

\usepackage{microtype}
\usepackage{graphicx}
\usepackage{subfigure}
\usepackage{booktabs} 

\usepackage{hyperref}

\usepackage[preprint]{icml2026}

\usepackage{amsmath}
\usepackage{amssymb}
\usepackage{mathtools}
\usepackage{amsthm}

\usepackage[capitalize,noabbrev]{cleveref}

\theoremstyle{plain}

\theoremstyle{definition}

\theoremstyle{remark}

\usepackage[textsize=tiny]{todonotes}

\icmltitlerunning{Signal-Adaptive Trust Regions for SNNs}

\begin{document}

\twocolumn[
\icmltitle{Signal-Adaptive Trust Regions for Gradient-Free Optimization of Recurrent Spiking Neural Networks}




\icmlsetsymbol{equal}{*}

\begin{icmlauthorlist}
\icmlauthor{Jinhao Li}{equal,sapient,thu}
\icmlauthor{Yuhao Sun}{equal,sapient}
\icmlauthor{Zhiyuan Ma}{thu}
\icmlauthor{Hao He}{thu}
\icmlauthor{Xinche Zhang}{thu}
\icmlauthor{Xing Chen}{sapient}
\icmlauthor{Jin Li}{sapient}
\icmlauthor{Sen Song}{thu}
\end{icmlauthorlist}

\icmlaffiliation{sapient}{Sapient Intelligence}
\icmlaffiliation{thu}{Tsinghua University}

\icmlcorrespondingauthor{Sen Song}{songsen@tsinghua.edu.cn}
\icmlcorrespondingauthor{Jin Li}{jin@sapient.inc\& electrixoul@outlook.com}

\icmlkeywords{Machine Learning, ICML}

\vskip 0.3in
]



\printAffiliationsAndNotice{\icmlEqualContribution} 

\newif\ifshowsuggest
\showsuggesttrue 

\newcommand{\syh}[1]{
    \ifshowsuggest
        \textcolor{red}{[SYH: #1]}
    \fi
}
\newif\ifshowsuggest
\showsuggesttrue 

\newcommand{\ljh}[1]{
    \ifshowsuggest
        \textcolor{blue}{[LJH: #1]}
    \fi
}

\begin{abstract}
Recurrent spiking neural networks (RSNNs) are a promising substrate for energy-efficient control policies, but training them for high-dimensional, long-horizon reinforcement learning remains challenging.
Population-based, gradient-free optimization circumvents backpropagation through non-differentiable spike dynamics by estimating gradients.
However, with finite populations, high variance of these estimates can induce harmful and overly aggressive update steps. Inspired by trust-region methods in reinforcement learning that constrain policy updates in distribution space, we propose \textbf{Signal-Adaptive Trust Regions (SATR)}, a distributional update rule that constrains relative change by bounding KL divergence normalized by an estimated signal energy. 
SATR automatically expands the trust region under strong signals and contracts it when updates are noise-dominated.
We instantiate SATR for Bernoulli connectivity distributions, which have shown strong empirical performance for RSNN optimization.
Across a suite of high-dimensional continuous-control benchmarks, SATR improves stability under limited populations and reaches competitive returns against strong baselines including PPO-LSTM.
In addition, to make SATR practical at scale, we introduce a bitset implementation for binary spiking and binary weights, substantially reducing wall-clock training time and enabling fast RSNN policy search. 
\end{abstract}

\section{Introduction}

Reinforcement Learning (RL) has achieved strong performance on high-dimensional continuous-control problems, enabling agents to acquire complex behaviors ranging from locomotion to manipulation~\citep{schulman2017proximal, lillicrap2015continuous}.
However, these successes typically rely on computationally intensive artificial neural networks (ANNs), which can limit scalable deployment on power and latency constrained edge platforms~\citep{hua2023edge}.

\begin{figure}[H]
  \centering 
  \includegraphics[width=\columnwidth]{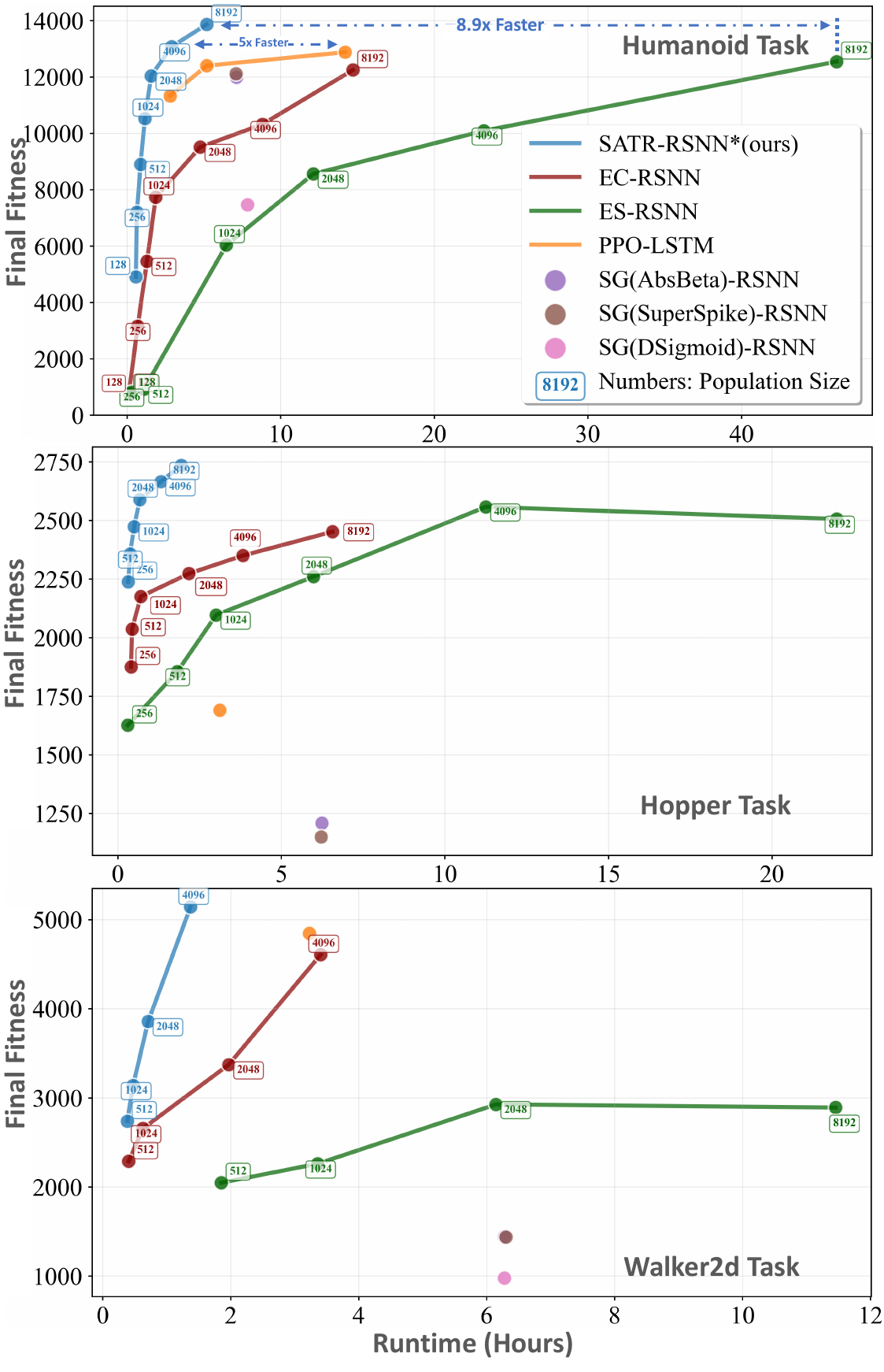}
  \caption{\textbf{Reward--runtime trade-off for RSNN policy optimization.} Final episodic return versus end-to-end wall-clock training time on \textit{Humanoid}, \textit{Hopper}, and \textit{Walker2d}. For population-based methods, each marker corresponds to one population size, and lines connect settings within the same method. Our approach ($\text{SATR-RSNN}^*$) achieves a more favorable reward--runtime trade-off, particularly in the limited-population regime.}
  \label{fig:reward_runtime}
\end{figure}

Spiking neural networks (SNNs) provide a compelling alternative: their event-driven dynamics enable sparse computation and can deliver substantial energy efficiency, particularly on neuromorphic hardware~\citep{roy2019towards, pei2019towards, Loihi2, davies2021advancing, liu2021low, modha2023neural, chen2024fully, shen2023learning, shrestha2022survey}.
Despite this promise, recurrent spiking policies have historically struggled to match ANN performance on high-dimensional, long-horizon control \citep{tavanaei2019deep, zanatta2024exploring, xu2025proxy, akl2023toward, wu2022training}.
Figure~\ref{fig:reward_runtime} previews our main empirical finding: a recurrent spiking RL agent that attains strong returns with substantially improved wall-clock efficiency and stability under limited population budgets.

A major obstacle for recurrent SNN control is reliable credit assignment through non-differentiable spike dynamics \citep{neftci2019surrogate, guo2023direct}.
The dominant paradigm, Backpropagation Through Time (BPTT) with surrogate gradients~\citep{werbos2002backpropagation}, and the memory cost of long unrolls often necessitates truncation, weakening the long-range temporal credit assignment required by complex control tasks \citep{neftci2019surrogate, bellec2018long, xiao2022online, zhu2022training, yin2023accurate}.
These challenges motivate gradient-free alternatives, such as Evolutionary Strategies (ES) and related population-based methods, which circumvent differentiability constraints by optimizing policies through population sampling \citep{salimans2017evolution}.
Among such approaches, Evolving Connectivity (EC) has emerged as a particularly effective framework for recurrent spiking policies \citep{wang2023evolving}.
EC models synapses as stochastic binary variables and optimizes a distribution over connectivity using Bernoulli parameters, and has demonstrated strong empirical performance on challenging high-dimensional control benchmarks, including high degree-of-freedom (DoF) tasks such as Humanoid \citep{wang2023evolving}.

At the same time, a key practical difficulty is shared by many population based optimizers: with finite populations, the resulting update estimates can be noisy, and training can become fragile when population budgets are limited \citep{salimans2017evolution, lehman2018safe, choromanski2019complexity}.
We argue that this fragility is not solely a matter of estimator variance; it is also a consequence of uncontrolled update step.
Population-based methods update a sampling distribution over parameters, and under finite-sample noise, a single iteration can induce an outsized change in distribution space, leading to abrupt behavioral shifts and destabilizing learning.
This issue is amplified in low-entropy regimes that are central to efficient spiking control---near-deterministic policies and especially sparse Bernoulli connectivity---where the curvature of distribution space is high.
In particular, for Bernoulli variables, as probabilities approach the boundaries ($0$ or $1$), small changes in the usual parameterization can correspond to disproportionately large changes in KL divergence \citep{amari1998natural, amari2016information, nielsen2020elementary}.

A natural way to control such displacement is to bound distributional change using information-theoretic trust-region ideas \citep{schulman2015trust, abdolmaleki2018maximum}.
However, fixed trust-region budgets implicitly assume that the underlying update direction is uniformly reliable.
This assumption breaks in population-based RL, where the reliability of the Monte Carlo update varies drastically with population size, task stochasticity, and training phase.
Consequently, a fixed KL budget can be overly conservative when the signal is strong, yet dangerously permissive when the signal is weak or noise-dominated---precisely the vulnerable regime where limited populations are most brittle.

We introduce \textbf{Signal-Adaptive Trust Regions (SATR)}, a population based update rule that couples the allowable KL divergence between successive sampling distributions to an estimate of signal strength.
Concretely, SATR bounds KL divergence normalized by an estimated signal energy, the squared norm of the population gradient estimate, expanding the trust region when the optimization signal is strong and contracting it when updates are noise-dominated.
We instantiate SATR for Bernoulli connectivity distributions in EC-trained RSNNs and evaluate it on a suite of high DoF continuous control benchmarks, where it improves stability under limited population budgets and achieves competitive returns against strong baselines including PPO-LSTM.
To complement algorithmic stability with practical throughput, we also introduce a bitset implementation for binary spikes and binary weights that substantially reduces wall-clock training time. 
Our contributions can be summarized as follows:
\begin{itemize}
    \item \textbf{Signal-adaptive trust regions for population-based RL.} We propose SATR, a novel distributional update rule that controls the relative change between successive sampling distributions by bounding KL divergence normalized by an empirical signal energy, automatically expanding under strong signals and contracting when updates are noise-dominated.
    \item \textbf{High performance on high-dimensional continuous control.} Across a diverse suite of long-horizon continuous-control benchmarks, we show that SATR improves training stability when population sizes are small and reaches returns competitive with strong baselines, including PPO-LSTM, while preserving the advantages of population-based RSNN optimization.
    \item \textbf{Bitset acceleration for binary spikes and binary weights.} We introduce a highly optimized bitset implementation for binary spiking and binary weights that significantly reduces wall-clock training time, enabling faster RSNN policy search and improving the reward--runtime trade-off.
\end{itemize}

\section{Problem Setup and Preliminaries}
\label{sec:prelim}

\subsection{RL objective and RSNN policies}
We consider finite-horizon episodic reinforcement learning in an MDP with observations $o_t$, actions $a_t$, and rewards $r_t$.
A recurrent spiking policy $\pi_\theta(a_t \mid o_{0:t})$ is parameterized by network parameters $\theta$ and is evaluated by rolling out an episode $\tau$ to obtain the (undiscounted) return
\begin{equation}
R(\tau)=\sum_{t=0}^{T-1} r_t \qquad (\gamma=1).
\end{equation}
Since our setting is inherently finite-horizon and we do not differentiate through time, using $\gamma=1$ is natural and avoids introducing an additional discount hyperparameter; all reported results are evaluated under a unified undiscounted evaluation protocol (Appendix~\ref{app:train-vs-eval}).
We write the expected return under parameters $\theta$ as
\begin{equation}
R(\theta)=\mathbb{E}_{\tau\sim \pi_\theta}[R(\tau)].
\end{equation}

We use an LIF-based recurrent SNN policy with a recurrent spiking layer and linear read-in/read-out connections; full neuron and synapse dynamics are provided in the Appendix.
Crucially, our method treats $\pi_\theta$ entirely as a black box: it requires only sampled returns and does not rely on backpropagation through spike dynamics.

\subsection{Population-based policy search}
To avoid differentiating through non-differentiable spike dynamics, we optimize a parameterized distribution through sampling.
Let $p_{\rho}(\theta)$ be a parametric distribution with parameters $\rho$.
The distributional objective is
\begin{equation}
J(\rho)=\mathbb{E}_{\theta\sim p_\rho}\!\left[R(\theta)\right].
\label{eq:dist_objective}
\end{equation}
A score-function identity yields
\begin{equation}
\nabla_{\rho}J(\rho)
=
\mathbb{E}_{\theta\sim p_\rho}\!\left[ R(\theta)\nabla_{\rho}\log p_\rho(\theta)\right].
\label{eq:score_grad}
\end{equation}

In practice, we estimate update directions from a finite population $\{\theta^{(n)}\}_{n=1}^N$ sampled from $p_\rho$.
To reduce sensitivity to reward scale and improve robustness, we apply a centered-rank fitness shaping to returns~\citep{wierstra2014natural, salimans2017evolution}.
Let $\mathrm{rank}(R(\theta^{(n)}))\in\{1,\dots,N\}$ denote the rank of $R(\theta^{(n)})$ within the population (ties handled by average rank), and define
\begin{equation}
\tilde{R}^{(n)}
=
\frac{\mathrm{rank}(R(\theta^{(n)})) - 1}{N-1} - \frac{1}{2},
\label{eq:centered_rank}
\end{equation}
so that $\tilde{R}^{(n)}\in[-\tfrac{1}{2},\tfrac{1}{2}]$ and is approximately zero-mean across the population.
We then subsequently form the Monte Carlo estimate as
\begin{equation}
\hat{g}
=
\frac{1}{N}\sum_{n=1}^{N}\tilde{R}^{(n)} \nabla_{\rho}\log p_\rho(\theta^{(n)}),
\label{eq:pop_grad_est}
\end{equation}
and update $\rho$ using $\hat{g}$.
In this paper, we refer to $\hat{g}$ as the population gradient estimate.

\subsection{Evolving Connectivity with Bernoulli distributions}
Following Evolving Connectivity (EC)~\citep{wang2023evolving}, we optimize a distribution over binary connectivity variables.
For each synapse we introduce independent binary random variables $\theta_i\in\{0,1\}$ with Bernoulli parameters $\rho_i\in(0,1)$ and assume a factorized sampling distribution
\begin{equation}
p_\rho(\theta)=\prod_{i=1}^{d}\mathrm{Bern}(\theta_i;\rho_i).
\label{eq:def_bern}
\end{equation}
For factorized Bernoulli variables,
\begin{equation}
\nabla_{\rho_i}\log p_\rho(\theta)
=
\frac{\theta_i-\rho_i}{\rho_i(1-\rho_i)}.
\label{eq:bern_score}
\end{equation}
EC applies a natural-gradient scaling, which multiplying by the inverse Fisher Information matrix for Bernoulli variables, yielding the natural-gradient direction
\begin{equation}
\tilde{\nabla}_{\rho_i} J(\rho)
=
\mathbb{E}_{\theta\sim p_\rho}\!\left[\tilde{R}(\theta)\,(\theta_i-\rho_i)\right],
\label{eq:ec_natgrad}
\end{equation}
where $\tilde{R}(\theta)$ denotes the centered-rank transformed return.

\subsection{Trust regions and KL divergence in reinforcement learning}
Trust-region methods in RL explicitly control the size of policy updates to avoid destructive changes in behavior~\citep{schulman2015trust, abdolmaleki2018maximum}.
A standard way to measure update size is the Kullback--Leibler (KL) divergence between the old and new policies, which is invariant to parameterization and closely related to the Fisher information metric.
TRPO solves a constrained first-order improvement problem: it maximizes a linear approximation of the objective subject to a KL trust-region constraint. Under the standard second-order approximation of the KL, this yields the classical normalized natural-gradient step:
\begin{equation}
\Delta\rho
=
\alpha \, F(\rho)^{-1} g,
\qquad
\alpha
=
\sqrt{\frac{2\delta}{g^\top F(\rho)^{-1} g}}.
\label{eq:trpo}
\end{equation}

where $g$ is the (Euclidean) gradient of the local surrogate objective w.r.t. $\rho$, $F(\rho)$ is the Fisher information matrix, and $\delta>0$ is the (scalar) KL budget.
Because the constraint introduces a single Lagrange multiplier, the resulting step length is necessarily a scalar normalization factor.

\section{Methods}
\label{sec:methods}

\subsection{Overview}
This section introduces \emph{Signal-Adaptive Trust Regions (SATR)}, a distribution-space trust-region rule for the population-based updates described in Sec.~\ref{sec:prelim}.
We assume a parameterized sampling distribution $p_\rho(\theta)$ over binary connectivity $\theta$, and a population estimate of the update direction $g$ obtained from episodic returns via centered-rank normalization.
SATR selects the distribution update $\Delta\rho$ by regulating the KL divergence between successive sampling distributions and scaling the allowable KL change with the \emph{signal energy} $\|g\|_2^2$.

We first formalize SATR at the level of general distributions (Sec.~\ref{subsec:satr_def}).
We then specialize SATR to the factorized Bernoulli family used in Evolving Connectivity, obtaining a closed-form update that we call \textbf{SATR-EC} and explaining why it is stable in both low-signal and near-boundary regimes (Sec.~\ref{subsec:satr_ec}).
Finally, we describe a bitset implementation that accelerates RSNN rollouts by exploiting binary spikes and binary connectivity/weights (Sec.~\ref{subsec:bitset_impl}).

\subsection{Signal-Adaptive Trust Regions (SATR)}
\label{subsec:satr_def}

We operate in a distributional optimization view of population-based RL (Sec.~\ref{sec:prelim}), where each iteration updates the parameters $\rho$ of a sampling distribution $p_\rho(\theta)$ rather than updating a single parameter vector.
A natural notion of update size is therefore the change in distribution space.
Following trust-region RL methods such as TRPO/MPO, we measure distributional change using KL divergence:
\begin{equation}
D_{\mathrm{KL}}\!\left(p_\rho \,\|\, p_{\rho'}\right)
\;\triangleq\;
\mathbb{E}_{\theta\sim p_\rho}\!\left[
\log\frac{p_\rho(\theta)}{p_{\rho'}(\theta)}
\right].
\label{eq:kl_def_general}
\end{equation}
Bounding $D_{\mathrm{KL}}(p_\rho \,\|\, p_{\rho'})$ constrains how far the sampling distribution can move per iteration, which is particularly important when updates are estimated from finite populations.

\paragraph{Signal Energy.}
Let $g$ denote the population update estimate for $\rho$ (Sec.~\ref{sec:prelim}); in our setting $g$ is computed from sampled connectivities and centered-rank normalized returns.
Because $g$ uses a finite population, its reliability varies with population size, task stochasticity, and training phase.
We quantify signal strength by the \emph{signal energy}
\begin{equation}
E \triangleq \|g\|_2^2,
\label{eq:signal_energy_def}
\end{equation}
Intuitively, when sampled perturbations produce a coherent update direction, $g$ tends to have a larger norm; when the update is noise-dominated, cancellations drive $\|g\|$ down.
Moreover, using centered-rank returns makes $g$ insensitive to the raw reward scale, so $\|g\|$ primarily reflects cross-sample agreement rather than magnitude of rewards.

\paragraph{Definition of SATR.}

A single global KL budget introduces only one constraint and therefore yields a single Lagrange multiplier, implying a scalar step-length (as in TRPO, Eq.~\eqref{eq:trpo}). In contrast, our sampling distribution is factorized across dimensions (Eq.~\ref{eq:def_bern}), so the KL decomposes across coordinates.
This enables a stronger and more appropriate element-wise trust-region design that (i) produces an element-wise step-length vector and (ii) prevents any single coordinate from causing an outsized distributional jump.

Assume a factorized family $p_\rho(\theta)=\prod_{i=1}^d p_{\rho_i}(\theta_i)$ so
that the KL decomposes as
$D_{\mathrm{KL}}(p_\rho\Vert p_{\rho'})=\sum_i D_i(\rho_i,\rho_i')$,
where $D_i(\rho_i,\rho_i')\triangleq D_{\mathrm{KL}}(p_{\rho_i}\Vert p_{\rho_i'})$.
We define SATR via the following element-wise signal-adaptive trust region:
\begin{align}
\max_{\Delta\rho\in\mathbb{R}^d}\; g^\top \Delta\rho
\quad
\text{s.t.}\quad
D_{KL}(\rho_i ||\rho_i+\Delta\rho_i)\;\le\;\delta\, g_i^2,\;\forall i
\label{eq:satr_def}
\end{align}
where $\delta>0$ is a constant. Consequently, when the population signal is strong, the feasible KL radius expands; when the signal is weak, the trust region contracts automatically, preventing noise-driven jumps in distribution space.

\paragraph{Local KL model and Fisher geometry.}
For small $\Delta\rho_i$, each one-dimensional KL admits the standard second-order
expansion in terms of the scalar Fisher information $F_i(\rho_i)$:
\begin{equation}
D_i(\rho_i,\rho_i+\Delta\rho_i)
=
\frac{1}{2}F_i(\rho_i)\,(\Delta\rho_i)^2
+
o\!\left((\Delta\rho_i)^2\right).
\label{eq:kl_1d_fisher}
\end{equation}
Substituting \eqref{eq:kl_1d_fisher} into the constraints of \eqref{eq:satr_def} yields
a separable quadratic constraint:
\begin{equation}
\frac{1}{2}F_i(\rho_i)(\Delta\rho_i)^2 \;\le\; \delta\, g_i^2,\qquad \forall i.
\label{eq:satr_quad_coord}
\end{equation}
Because the objective also separates as $\sum_i g_i\Delta\rho_i$, the optimization
decouples across dimensions. For any $i$ with $g_i\neq 0$, the optimum lies on the
constraint boundary with $\mathrm{sign}(\Delta\rho_i)=\mathrm{sign}(g_i)$, giving the
closed-form solution
\begin{equation}
\Delta\rho_i^\star
= \alpha_ig_i =
\sqrt{\frac{2\delta}{F_i(\rho_i)}}\, g_i,
\label{eq:satr_solution_general}
\end{equation}
Since the distribution is assumed to be factorized, $F$ is diagonal. Therefore, we could reform it into the form of vectorized trust region definition.
Thus, SATR naturally yields an element-wise step-length vector $\alpha$ rather than a simple scalar normalization factor.

\subsection{SATR-EC: SATR specialized to Bernoulli connectivity}
\label{subsec:satr_ec}

\begin{algorithm}[t]
  \caption{SATR-EC for RSNN policy optimization}
  \label{alg:satr_ec}
  \begin{algorithmic}[1]
    \STATE {\bfseries Input:} initial $\rho\in(0,1)^d$, population size $N$, stepsize $\eta$, environment $\mathcal{E}$
    \REPEAT
      \STATE Sample $\theta^{(1)},\dots,\theta^{(N)} \sim \prod_i \mathrm{Bern}(\theta_i;\rho_i)$
      \FOR{\textbf{(in parallel)} $n=1$ {\bfseries to} $N$}
        \STATE Roll out one episode with RSNN policy instantiated by $\theta^{(n)}$
        \STATE Record return $R(\theta^{(n)})$
      \ENDFOR
      \STATE Compute centered-rank returns $\tilde{R}^{(1)},\dots,\tilde{R}^{(N)}$
      \STATE $g \leftarrow \frac{1}{N}\sum_{n=1}^N \tilde{R}^{(n)}(\theta^{(n)}-\rho)$
      \STATE $\rho \leftarrow \rho + \eta\,\sqrt{\rho\odot(1-\rho)}\odot g$
    \UNTIL{converged}
  \end{algorithmic}
\end{algorithm}

We now instantiate SATR for the factorized Bernoulli family used by Evolving Connectivity (Sec.~\ref{sec:prelim}).
Recall that $p_\rho(\theta)=\prod_{i=1}^{d}\mathrm{Bern}(\theta_i;\rho_i)$ with $\rho\in(0,1)^d$, and that EC provides a population update estimate
$g\in\mathbb{R}^d$ from centered-rank normalized returns.
The Bernoulli family has a simple information geometry.
For a single Bernoulli mean parameter $\rho_i$, the Fisher information is
\begin{equation}
F_i(\rho_i)=\frac{1}{\rho_i(1-\rho_i)},
\label{eq:bern_fisher_diag}
\end{equation}
and for the factorized family the Fisher matrix is diagonal,
$F(\rho)=\mathrm{diag}(F_1(\rho_1),\dots,F_d(\rho_d))$.
Consequently, the KL divergence admits the standard local expansion
\begin{align}
D_{\mathrm{KL}}\!\left(p_\rho \,\|\, p_{\rho+\Delta\rho}\right)
&=
\frac{1}{2}\Delta\rho^\top F(\rho)\Delta\rho
+
o(\|\Delta\rho\|^2) \\
\;&\approx\;
\frac{1}{2}\sum_{i=1}^{d}\frac{(\Delta\rho_i)^2}{\rho_i(1-\rho_i)}.
\label{eq:bern_local_kl}
\end{align}
This expression makes the boundary curvature explicit: as $\rho_i\to 0$ or $1$, the denominator $\rho_i(1-\rho_i)$ vanishes and the same Euclidean step $\Delta\rho_i$ corresponds to a much larger displacement in distribution space.

Applying the element-wise SATR constraints \eqref{eq:satr_def} under the local model \eqref{eq:bern_local_kl} yields, for each coordinate,
\begin{equation}
\Delta\rho = \eta\;F(\rho)^{-1/2}g = \eta\;\sqrt{\rho\odot(1-\rho)}\;\odot\; g.
\label{eq:satr_ec_update}
\end{equation}
where we define the learning rate as $\eta = \sqrt{2\delta}$, yielding the practical update used in implementation.
We refer to Eq.~\ref{eq:satr_ec_update} as \textbf{SATR-EC} (SATR for EC).
The overall procedure is summarized in Algorithm~\ref{alg:satr_ec}.

\paragraph{Signal-adaptive KL property.}
SATR-EC satisfies the SATR principle in a particularly transparent way.
Substituting $\Delta\rho_i=\eta\sqrt{\rho_i(1-\rho_i)}\,g_i$ into Eq.~\ref{eq:bern_local_kl} gives
\begin{equation}
D_{\mathrm{KL}}\!\left(p_\rho \,\|\, p_{\rho+\Delta\rho}\right)
\;\approx\;
\frac{1}{2}\sum_{i=1}^{d}
\frac{\eta^2\rho_i(1-\rho_i)g_i^2}{\rho_i(1-\rho_i)}
\;=\;
\frac{\eta^2}{2}\|g\|_2^2.
\label{eq:satr_ec_kl_property}
\end{equation}
Thus the normalized displacement
$
D_{\mathrm{KL}}(p_\rho \,\|\, p_{\rho+\Delta\rho})/\|g\|_2^2
$
is approximately constant (equal to $\eta^2/2$): the realized trust region expands when $\|g\|$ is large and contracts automatically when $\|g\|$ is small.

\paragraph{Why SATR-EC is stable.}
Eq.~\ref{eq:satr_ec_kl_property} directly captures the stability mechanism targeted by SATR.
(i) under noisy finite-population updates, gradient cancellations reduce $\|g\|_2$, thereby shrinking the KL step automatically; (ii) SATR-EC is boundary-aware because $\sqrt{\rho_i(1-\rho_i)}\to 0$ as $\rho_i\to 0$ or $1$, preventing curvature-driven KL blow-ups near deterministic connectivity.

To highlight the contrast, applying the baseline EC step $\Delta\rho=\eta g$ to Eq.~\ref{eq:bern_local_kl} yields
\begin{equation}
D_{\mathrm{KL}}\!\left(p_\rho \,\|\, p_{\rho+\eta g}\right)
\;\approx\;
\frac{\eta^2}{2}\sum_{i=1}^{d}\frac{g_i^2}{\rho_i(1-\rho_i)},
\label{eq:ec_kl_blowup}
\end{equation}
which can become arbitrarily large as any $\rho_i$ approaches a boundary.
SATR-EC removes this curvature amplification by equalizing the local KL geometry through $\sqrt{\rho_i(1-\rho_i)}$.

\paragraph{Practical update.}
We update $\rho\leftarrow\rho+\Delta\rho$ using Eq.~\ref{eq:satr_ec_update}.
To maintain $\rho\in(0,1)$ we optionally apply a tiny numerical clamp
$\rho\leftarrow\mathrm{clip}(\rho,\epsilon,1-\epsilon)$ (e.g., $\epsilon=10^{-3}$),
which is employed only for basic numerical safety and does not play the role of a stability heuristic.

\subsection{Bitset acceleration for binary RSNN rollouts}
\label{subsec:bitset_impl}


To improve rollout throughput, we exploit that (i) the sampled connectivity $\theta$ in EC and (ii) RSNN spiking activity are binary.
Moreover, a sampled connectivity $\theta$ remains fixed throughout an episode.
Instead of representing these objects as byte or floating-point arrays and performing dense matrix multiplications, we pack binary vectors and matrices into machine words (bitsets).
This reduces memory traffic and enables synaptic integration to be computed using hardware-supported bitwise operations.

Consider a presynaptic spike vector $s_t\in\{0,1\}^{d_{\mathrm{in}}}$ at timestep $t$ and a binary connectivity mask $m_j\in\{0,1\}^{d_{\mathrm{in}}}$ for postsynaptic neuron $j$.
Let $w$ denote the word size (e.g., $w=64$) and $B=\lceil d_{\mathrm{in}}/w\rceil$ the number of packed words.
We pack $s_t$ into words $\{S_{t,b}\}_{b=1}^B$ and $m_j$ into words $\{M_{j,b}\}_{b=1}^B$.
The masked spike count (the binary dot product) can then be computed as
\begin{equation}
m_j^\top s_t
\;=\;
\sum_{b=1}^{B}
\mathrm{popcount}\!\left(M_{j,b}\ \&\ S_{t,b}\right),
\label{eq:bitset_and_popcount}
\end{equation}
where $\&$ denotes bitwise \texttt{AND} and \texttt{popcount} returns the number of set bits in a word.
When synaptic weights are binary, Eq.~\ref{eq:bitset_and_popcount} replaces the dominant multiply--accumulate cost of synaptic integration with a small number of word-level bit operations and integer additions.

The complexity of Eq.~\ref{eq:bitset_and_popcount} scales as $O(B)$ word operations per postsynaptic neuron, i.e., $O(d_{\mathrm{in}}/w)$, and it benefits from highly optimized bitwise instructions on modern CPUs/GPUs.
Because $\theta$ is fixed within an episode, the packed connectivity words $\{M_{j,b}\}$ are constructed once per episode, while spikes are packed on the fly at each timestep.
In practice, this significantly improves cache locality and reduces memory bandwidth pressure compared to floating-point matrix multiplications, which is especially beneficial when evaluating many population members in parallel.

The bitset implementation is purely computational: it does not change the objective, estimator, or SATR-EC update rule.
Its role is to significantly increase parallel rollout throughput and reduce wall-clock time per population evaluation, making limited-population training regimes and large benchmark suites computationally practical.

\section{Experimental Setup}
We evaluate the proposed SATR-RSNN on standard continuous control benchmarks provided by the Brax physics engine, specifically the \texttt{humanoid(17-Dof)}, \texttt{hopper(3-Dof)}, and \texttt{walker2d(6-Dof)} environments. We compare our method against 3 primary baselines:
\begin{itemize}
    \item \textbf{PPO-LSTM:} A standard Proximal Policy Optimization agent with LSTM recurrence, representing the state-of-the-art in gradient-based RL.
    \item \textbf{ES-RSNN:} An RSNN trained via Evolutionary Strategies~\citep{salimans2017evolution}, representing a gradient-free biological baseline.
    \item \textbf{EC-RSNN:} An RSNN trained with the original Evolving Connectivity~\citep{wang2023evolving}, serving as an ablation baseline.
    \item \textbf{SG-RSNN:} An RSNN trained in PPO with the surrogate gradient method, which is specially designed for spiking neuron networks. We implemented three different surrogate gradient functions: AbsBeta\cite{Esser2016ConvolutionalNF}, SuperSpike\cite{SuperSpike}, and $\mathrm{Sigmoid}^{\prime}$ (DSigmoid) \cite{zenke2021remarkable}.
\end{itemize}

All recurrent policies (RSNNs and the LSTM baseline) use a single recurrent hidden layer and are configured to have comparable numbers of trainable parameters across methods.
Full detailed description of all networks and training parameters is in Appendix~\ref{app:ppo}.

\paragraph{Training budget and evaluation protocol.}

All methods are implemented in JAX~\citep{jax2018github} and evaluated using the Brax~\citep{freeman2021brax} physics engine on the same hardware, ensuring a strictly consistent execution backend and fully fair wall-clock comparison (see Appendix~\ref{app:train-vs-eval} for the unified evaluation metric and protocol). Results are reported as mean $\pm$ standard deviation over \textbf{3 random seeds} in Table~\ref{tab:all_performance}. 

For population-based methods (ES, EC, and SATR), training proceeds in generations. We run 2,000 generations on Humanoid and 1,000 generations on Hopper and Walker2d. For gradient-based baselines, the training budget is controlled by the number of gradient updates.
We consider two training settings for gradient-based baselines. In the non-accelerated setting, PPO is trained for 2.2 million gradient steps and surrogate-gradient (SG) baselines are trained for 250k gradient steps on each task. This setting serves as the comparison against the non-accelerated SATR.
To enable fair wall-clock comparisons under identical rollout execution with bitset acceleration, PPO is trained for 500k and 850k gradient steps to match the runtime of SATR with population sizes of 4096 and 8192, respectively.
This two-setting design disentangles optimization budget from system-level acceleration. It enables fair comparisons in terms of training budget (full PPO baseline) and wall-clock runtime under acceleration (runtime-matched settings), ensuring that any additional speedup observed for SATR reflects system-level acceleration rather than reduced optimization effort.

\begin{table*}[t]
\centering
\caption{Performance Summary on Different Tasks. }
\resizebox{\textwidth}{!}{
\begin{tabular}{c|ccc|ccc|ccc}
\toprule
Pop 
& \multicolumn{3}{c}{Humanoid (17-Dof)} 
& \multicolumn{3}{c}{Hopper (3-Dof)} 
& \multicolumn{3}{c}{Walker2d (6-Dof)} \\
\cmidrule(lr){2-4} \cmidrule(lr){5-7} \cmidrule(lr){8-10}
 & ES & EC & \textbf{SATR(Ours) }
 & ES & EC & \textbf{SATR(Ours) }
 & ES & EC & \textbf{SATR(Ours) }\\
\midrule
8192 & 12544 ± 1175 & 12242 ± 1903 & $\mathbf{13860 \pm 1107}$
     & 2505 ± 160  & 2451 ± 463  & $\mathbf{2735 \pm 160}$
     & -- & -- & -- \\
4096 & 10087 ± 1162 & 10319 ± 1196 & $\mathbf{13072 \pm 941}$
     & 2557 ± 294  & 2350 ± 405  & $\mathbf{2665 \pm 188}$
     & 2891 ± 167 & 4605 ± 148 & $\mathbf{5143 \pm 459} $\\
2048 & 8558 ± 659 & 9506 ± 939 & $\mathbf{12037 \pm 278}$
     & 2260 ± 54 & 2273 ± 677 & $\mathbf{2588 \pm 79}$
     & 2925 ± 295 & 3372 ± 721 & $\mathbf{3856 \pm 607}$ \\
1024 & 6031 ± 993 & 7729 ± 1028 & $\mathbf{10520 \pm 457}$
     & 2096 ± 66 & 2175 ± 755 & $\mathbf{2473 \pm 237}$
     & 2259 ± 164 & 2656 ± 509 & $\mathbf{3139 \pm 806}$ \\
512  & 922 ± 16 & 5462 ± 434 & $\mathbf{8928 \pm 119}$
     & 1855 ± 56 & 2036 ± 329 & $\mathbf{2356 \pm 138}$
     & 2047 ± 295 & 2288 ± 616 & $\mathbf{2738 \pm 298}$ \\
256  & 823 ± 24 & 3155 ± 2508 & $\mathbf{6656 \pm 1643}$
     & 1625 ± 203 & 1875 ± 51 & $\mathbf{2238 \pm 97}$
     & -- & -- & -- \\
128  & 750 ± 42 & 765 ± 36 & $\mathbf{4908 \pm 32}$
     & -- & -- & --
     & -- & -- & -- \\
\bottomrule
\end{tabular}
}
\label{tab:all_performance}
\end{table*}

\begin{figure*}[ht] 
  \centering
  \includegraphics[width=\textwidth]{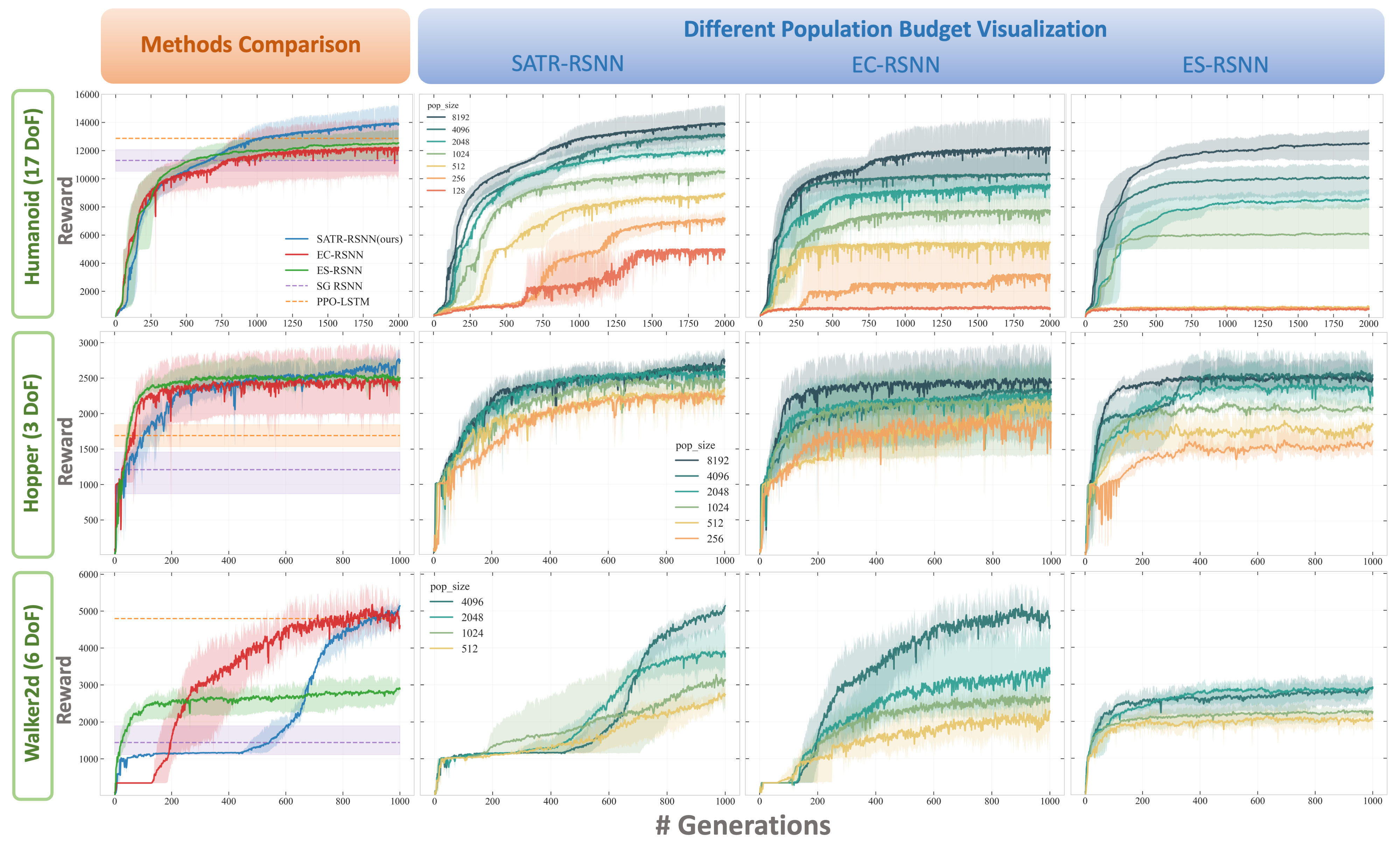}
  \caption{\textbf{Performance Benchmarks Across Tasks and Population Scales.} Learning curves for three continuous control tasks under varying population budgets ($N$). Each column represents a distinct population size, demonstrating the scalability of our approach. SATR (ours) consistently achieves the highest terminal returns and exhibits superior convergence stability compared to EC and ES.}
  \label{fig:exp_CurveOf3Tasks}
\end{figure*}

\section{Results}
\subsection{Performance Analysis}

We evaluate the proposed Signal-Adaptive Trust Region (SATR) on three Brax continuous-control benchmarks—Humanoid, Hopper, and Walker2d—and compare against population-based baselines (EC, ES) and gradient-based methods (SG, PPO).

\textbf{Overall performance.}
As shown in Table~\ref{tab:all_performance}, SATR consistently and significantly outperforms ES and standard EC across all tasks. On the most challenging Humanoid task with $N=8192$, SATR achieves a return of \textbf{13,860}. Learning curves in Figure~\ref{fig:exp_CurveOf3Tasks} further show that SATR converges faster and more stably than standard population-based methods across all benchmarks.

\textbf{Robustness to limited population budgets.}
To assess robustness under noisy updates, we reduce the population size from $N=8192$ to $N=128$. Table~\ref{tab:all_performance} shows a clear performance divergence as $N$ decreases, revealing the sensitivity of different methods to sampling noise. On Humanoid, ES drops from $12{,}544$ to $922$ and EC to $5{,}462$ at $N=512$, while SATR retains a substantial $8{,}928$ ($1.6\times$ over EC). At $N=128$, SATR still achieves $4{,}908$, whereas EC collapses to $765$. Similar trends are observed on Hopper and Walker2d. These results support the signal-adaptive trust-region design, which contracts updates under weak signals and avoids noise-induced distributional collapse.

\textbf{Comparison to gradient-based baselines.}
We further compare SATR with gradient-based baselines \citep{SuperSpike, Esser2016ConvolutionalNF, zenke2021remarkable}, including advanced surrogate-gradient (SG) RSNNs and PPO-LSTM. As shown in Figure~\ref{fig:exp_CurveOfSG}, SATR outperforms the best-performing SG-RSNN configuration on the Humanoid task, consistent with the difficulty of applying surrogate gradients and truncated backpropagation-through-time to long-horizon recurrent spiking policies.

Under a matched wall-clock training budget, SATR achieves comparable or higher final returns than PPO-LSTM across Humanoid, Hopper, and Walker2d, despite relying on binary spiking dynamics and gradient-free optimization. These results indicate that signal-adaptive, distributional updates provide a competitive and stable alternative to gradient-based training for recurrent spiking control.

\begin{table}[!ht]
\centering
\caption{Runtime (seconds) comparison between Binary RSNN($RSNN~w/o~bitset$) and Bitset Accelerated Binary RSNN($RSNN^*$) in humanoid task.}
\resizebox{\columnwidth}{!}{
    \begin{tabular}{lccccc}
    \hline
    Pop & 512 & 1024 & 2048 & 4096 & 8192 \\
    \hline
    $RSNN^*$ & 3168 & 4178 & 5645 & 10431 & 18668 \\
    $RSNN~w/o~bitset$ & 5212 & 6912 & 15569 & 26944 & 51098 \\
    $\times$ faster & 1.64 & 1.65 & 2.75 & 2.58 & 2.74 \\
    \hline
    \end{tabular}
}
\label{tab:ablation_acceleration}
\end{table}

\subsection{Computational Efficiency and Reward--Runtime Trade-off}

\paragraph{Compute matching and attribution.}
Under matched population size and generations, SATR evaluates the same number of rollouts as EC. Therefore, under the same non-accelerated rollout implementation, SATR does not yield an inherent wall-clock speedup over EC; any runtime improvement must come from system-level acceleration, rather than differences in rollout volume or evaluation frequency.

\paragraph{Bitset execution for binary RSNNs.}
To make SATR practical at scale, we introduce a high-performance bitset execution engine tailored to binary spikes and binary weights.
By replacing floating-point matrix multiplications with bitwise operations and \texttt{popcount}, it increases rollout throughput by up to $\sim$2.7$\times$ (Table~\ref{tab:ablation_acceleration}), which directly improves the reward--runtime trade-off in Fig.~\ref{fig:reward_runtime}.

Beyond final return, we evaluate computational efficiency in terms of end-to-end wall-clock training time.
Figure~\ref{fig:reward_runtime} summarizes the reward--runtime trade-off on Humanoid, Hopper, and Walker2d.
Across all tasks, SATR achieves higher final returns at the same or lower wall-clock time compared to population-based baselines such as ES and EC, with the advantage being most pronounced on the computationally demanding Humanoid task.
At matched performance levels, SATR requires significantly less runtime, enabling faster exploration of the population size--performance frontier.

These gains arise from system-level acceleration rather than changes to the learning objective or update rule.
Since connectivity samples remain fixed within each episode and spike activity is binary, bitset-based execution improves efficiency without altering the optimization algorithm.
In practice, on Humanoid, EC already provides an approximately 3.24 speedup over ES due to its more efficient population-based optimization. Building on this baseline, our bitset backend further reduces wall-clock training time by an additional 2.74×, lowering runtime from 51,098 s to 18,668 s. Together, these effects result in an overall speedup of up to 8.9× compared to ES. At matched control performance, SATR also achieves approximately 5× faster training than PPO-LSTM, yielding a favorable reward–runtime trade-off.

\paragraph{Energy efficiency estimation.} 

We additionally provide an analytical estimate of the on-chip energy (Loihi~\citep{Loihi2}), suggesting about 400$\times$--3200$\times$ lower energy consumption than PPO-LSTM on GPU, details in Appendix~\ref{app:power}. 
We emphasize that these RSNN numbers are analytical estimates based on published neuromorphic per-operation energy figures~\citep{Loihi2}, intended for order-of-magnitude comparison; direct measurements on neuromorphic hardware are left for future work.

\subsection{Ablation: Fixed vs.\ Signal-Adaptive Trust Regions}

To isolate the role of signal adaptivity, we compare SATR with a fixed-KL-budget trust-region variant added to the EC baseline (denoted EC+TR).  EC+TR imposes a TRPO-style KL constraint in the distribution-parameter space, which leads to a closed-form normalized natural-gradient step
\begin{equation}
\Delta \rho \leftarrow \sqrt{2\delta}\;\frac{g}{\sqrt{g^\top F^{-1} g}},
\label{eq:fixed_tr_update}
\end{equation}
where $g$ is the population update direction in distribution space and $F$ is the Bernoulli Fisher information matrix defining the local KL metric. 

Table~\ref{tab:ablation_tr} shows that a fixed KL budget can improve EC stability under small populations, but performance is sensitive to the choice of $\delta$ and does not transfer reliably across population sizes.
For example, at Pop=256, moderately sized budgets improve over EC, while overly large budgets can trigger collapse; at Pop=512 and 1024, the best-performing $\delta$ shifts substantially, and no single setting is strong across all regimes.
This sensitivity makes EC+TR difficult to use in practice when the population budget changes.

In contrast, SATR-EC achieves strong performance across all tested populations without tuning a fixed KL budget (Table~\ref{tab:ablation_tr}), consistently matching or outperforming the best EC+TR setting.
These results support our central hypothesis: scaling the effective KL radius with the reliability of the population signal yields a more robust stability--performance trade-off than enforcing any single fixed KL budget.

\begin{table}[!ht]
    \centering
    \caption{Ablation study of the difference trust region method in Humanoid task, TR means the traditional TRPO method. The \textbf{bold} number is the highest score, and the \underline{underline} score is the score between the score of EC baseline and the proposed SATR.}
    \resizebox{\columnwidth}{!}{
        \begin{tabular}{lcccc}
        \hline
        Method &$\delta \times param$ & Pop=256 & Pop=512 & Pop=1024 \\
        \hline
        EC & - & $3155\pm2508$ & $5462\pm434$ & $7729\pm1028$ \\
        EC+TR&3 & - & $3961\pm1658$ & $6832\pm580$ \\
        EC+TR&10 & - & \underline{$5614\pm529$} & $7510\pm1242$ \\
        EC+TR&30 & $2480\pm1800$ & $3738\pm1670$ & \underline{$8163\pm865$} \\
        EC+TR&100 & \underline{$4259\pm639$} & $3690\pm1617$ & \underline{$7855\pm980$} \\
        EC+TR&300 & \underline{$4682\pm263$} & \underline{$6592\pm1846$} & \underline{$8122\pm343$} \\
        EC+TR&1000 & $923\pm121$ & $4918\pm1691$ & $6812\pm1275$ \\
        $\mathbf{SATR}$& - & $\mathbf{6656\pm1643}$  & $\mathbf{8928\pm119}$ &  $\mathbf{10520\pm457}$\\
        \hline
        \end{tabular}
    }
    \label{tab:ablation_tr}
\end{table}

\section{Conclusion}
We propose Signal-Adaptive Trust Regions (SATR) for population-based learning, which scale the KL divergence between successive sampling distributions based on population update reliability measured via signal energy.
Specialized to Bernoulli connectivity and RSNNs, and combined with a bitset implementation exploiting binary spikes and connectivity, SATR consistently improves stability under small population budgets and delivers an improved reward–runtime trade-off across Brax continuous-control benchmarks, enabling  scalable RSNN policy search.


\section*{Impact Statement}
This paper presents work whose goal is to advance the field of Machine
Learning. There are many potential societal consequences of our work, none
which we feel must be specifically highlighted here.

\bibliography{example_paper}

@article{werbos2002backpropagation,
  title={Backpropagation through time: what it does and how to do it},
  author={Werbos, Paul J},
  journal={Proceedings of the IEEE},
  volume={78},
  number={10},
  pages={1550--1560},
  year={2002},
  publisher={IEEE}
}

@article{shrestha2022survey,
  title={A survey on neuromorphic computing: Models and hardware},
  author={Shrestha, Amar and Fang, Haowen and Mei, Zaidao and Rider, Daniel Patrick and Wu, Qing and Qiu, Qinru},
  journal={IEEE Circuits and Systems Magazine},
  volume={22},
  number={2},
  pages={6--35},
  year={2022},
  publisher={IEEE}
}

@article{shen2023learning,
  title={Learning the Plasticity: Plasticity-Driven Learning Framework in Spiking Neural Networks},
  author={Shen, Guobin and Zhao, Dongcheng and Dong, Yiting and Li, Yang and Zhao, Feifei and Zeng, Yi},
  journal={arXiv preprint arXiv:2308.12063},
  year={2023}
}

@article{chen2024fully,
  title={Fully spiking actor network with intralayer connections for reinforcement learning},
  author={Chen, Ding and Peng, Peixi and Huang, Tiejun and Tian, Yonghong},
  journal={IEEE Transactions on Neural Networks and Learning Systems},
  volume={36},
  number={2},
  pages={2881--2893},
  year={2024},
  publisher={IEEE}
}

@misc{jax2018github,
  author = {James Bradbury and Roy Frostig and Peter Hawkins and Matthew James Johnson and Chris Leary and Dougal Maclaurin and George Necula and Adam Paszke and Jake Vander{P}las and Skye Wanderman-{M}ilne and Qiao Zhang},
  title = {{JAX}: composable transformations of {P}ython+{N}um{P}y programs},
  url = {http://github.com/jax-ml/jax},
  note = {Version 0.3.13},
  year = {2018}
}

@article{schulman2017proximal,
  title={Proximal policy optimization algorithms},
  author={Schulman, John and Wolski, Filip and Dhariwal, Prafulla and Radford, Alec and Klimov, Oleg},
  journal={arXiv preprint arXiv:1707.06347},
  year={2017}
}

@article{lillicrap2015continuous,
  title={Continuous control with deep reinforcement learning},
  author={Lillicrap, Timothy P and Hunt, Jonathan J and Pritzel, Alexander and Heess, Nicolas and Erez, Tom and Tassa, Yuval and Silver, David and Wierstra, Daan},
  journal={arXiv preprint arXiv:1509.02971},
  year={2015}
}

@article{abdolmaleki2018maximum,
  title={Maximum a posteriori policy optimisation},
  author={Abdolmaleki, Abbas and Springenberg, Jost Tobias and Tassa, Yuval and Munos, Remi and Heess, Nicolas and Riedmiller, Martin},
  journal={arXiv preprint arXiv:1806.06920},
  year={2018}
}

@article{roy2019towards,
  title={Towards spike-based machine intelligence with neuromorphic computing},
  author={Roy, Kaushik and Jaiswal, Akhilesh and Panda, Priyadarshini},
  journal={Nature},
  volume={575},
  number={7784},
  pages={607--617},
  year={2019},
  publisher={Nature Publishing Group UK London}
}

@article{pei2019towards,
  title={Towards artificial general intelligence with hybrid Tianjic chip architecture},
  author={Pei, Jing and Deng, Lei and Song, Sen and Zhao, Mingguo and Zhang, Youhui and Wu, Shuang and Wang, Guanrui and Zou, Zhe and Wu, Zhenzhi and He, Wei and others},
  journal={Nature},
  volume={572},
  number={7767},
  pages={106--111},
  year={2019},
  publisher={Nature Publishing Group UK London}
}

@article{tavanaei2019deep,
  title={Deep learning in spiking neural networks},
  author={Tavanaei, Amirhossein and Ghodrati, Masoud and Kheradpisheh, Saeed Reza and Masquelier, Timoth{\'e}e and Maida, Anthony},
  journal={Neural networks},
  volume={111},
  pages={47--63},
  year={2019},
  publisher={Elsevier}
}

@article{zanatta2024exploring,
  title={Exploring spiking neural networks for deep reinforcement learning in robotic tasks},
  author={Zanatta, Luca and Barchi, Francesco and Manoni, Simone and Tolu, Silvia and Bartolini, Andrea and Acquaviva, Andrea},
  journal={Scientific Reports},
  volume={14},
  number={1},
  pages={30648},
  year={2024},
  publisher={Nature Publishing Group UK London}
}

@article{xu2025proxy,
  title={Proxy Target: Bridging the Gap Between Discrete Spiking Neural Networks and Continuous Control},
  author={Xu, Zijie and Bu, Tong and Hao, Zecheng and Ding, Jianhao and Yu, Zhaofei},
  journal={arXiv preprint arXiv:2505.24161},
  year={2025}
}

@article{Loihi2,
  title={Loihi: A neuromorphic manycore processor with on-chip learning},
  author={Davies, Mike and Srinivasa, Narayan and Lin, Tsung-Han and Chinya, Gautham and Cao, Yongqiang and Choday, Sri Harsha and Dimou, Georgios and Joshi, Prasad and Imam, Nabil and Jain, Shweta and others},
  journal={Ieee Micro},
  volume={38},
  number={1},
  pages={82--99},
  year={2018},
  publisher={IEEE}
}

@article{neftci2019surrogate,
  title={Surrogate gradient learning in spiking neural networks: Bringing the power of gradient-based optimization to spiking neural networks},
  author={Neftci, Emre O and Mostafa, Hesham and Zenke, Friedemann},
  journal={IEEE Signal Processing Magazine},
  volume={36},
  number={6},
  pages={51--63},
  year={2019},
  publisher={IEEE}
}

@article{bellec2018long,
  title={Long short-term memory and learning-to-learn in networks of spiking neurons},
  author={Bellec, Guillaume and Salaj, Darjan and Subramoney, Anand and Legenstein, Robert and Maass, Wolfgang},
  journal={Advances in neural information processing systems},
  volume={31},
  year={2018}
}

@article{guo2023direct,
  title={Direct learning-based deep spiking neural networks: a review},
  author={Guo, Yufei and Huang, Xuhui and Ma, Zhe},
  journal={Frontiers in Neuroscience},
  volume={17},
  pages={1209795},
  year={2023},
  publisher={Frontiers Media SA}
}

@article{salimans2017evolution,
  title={Evolution strategies as a scalable alternative to reinforcement learning},
  author={Salimans, Tim and Ho, Jonathan and Chen, Xi and Sidor, Szymon and Sutskever, Ilya},
  journal={arXiv preprint arXiv:1703.03864},
  year={2017}
}

@article{choromanski2019complexity,
  title={From complexity to simplicity: Adaptive es-active subspaces for blackbox optimization},
  author={Choromanski, Krzysztof M and Pacchiano, Aldo and Parker-Holder, Jack and Tang, Yunhao and Sindhwani, Vikas},
  journal={Advances in Neural Information Processing Systems},
  volume={32},
  year={2019}
}

@inproceedings{lehman2018safe,
  title={Safe mutations for deep and recurrent neural networks through output gradients},
  author={Lehman, Joel and Chen, Jay and Clune, Jeff and Stanley, Kenneth O},
  booktitle={Proceedings of the Genetic and Evolutionary Computation Conference},
  pages={117--124},
  year={2018}
}

@article{wierstra2014natural,
  title={Natural evolution strategies},
  author={Wierstra, Daan and Schaul, Tom and Glasmachers, Tobias and Sun, Yi and Peters, Jan and Schmidhuber, J{\"u}rgen},
  journal={The Journal of Machine Learning Research},
  volume={15},
  number={1},
  pages={949--980},
  year={2014},
  publisher={JMLR. org}
}

@article{wang2023evolving,
  title={Evolving connectivity for recurrent spiking neural networks},
  author={Wang, Guan and Sun, Yuhao and Cheng, Sijie and Song, Sen},
  journal={Advances in Neural Information Processing Systems},
  volume={36},
  pages={2991--3007},
  year={2023}
}

@article{amari1998natural,
  title={Natural gradient works efficiently in learning},
  author={Amari, Shun-Ichi},
  journal={Neural computation},
  volume={10},
  number={2},
  pages={251--276},
  year={1998},
  publisher={MIT Press}
}

@inproceedings{schulman2015trust,
  title={Trust region policy optimization},
  author={Schulman, John and Levine, Sergey and Abbeel, Pieter and Jordan, Michael and Moritz, Philipp},
  booktitle={International conference on machine learning},
  pages={1889--1897},
  year={2015},
  organization={PMLR}
}

@article{freeman2021brax,
  title={Brax--a differentiable physics engine for large scale rigid body simulation},
  author={Freeman, C Daniel and Frey, Erik and Raichuk, Anton and Girgin, Sertan and Mordatch, Igor and Bachem, Olivier},
  journal={arXiv preprint arXiv:2106.13281},
  year={2021}
}

@article{hua2023edge,
  title={Edge computing with artificial intelligence: A machine learning perspective},
  author={Hua, Haochen and Li, Yutong and Wang, Tonghe and Dong, Nanqing and Li, Wei and Cao, Junwei},
  journal={ACM Computing Surveys},
  volume={55},
  number={9},
  pages={1--35},
  year={2023},
  publisher={ACM New York, NY}
}

@article{davies2021advancing,
  title={Advancing neuromorphic computing with loihi: A survey of results and outlook},
  author={Davies, Mike and Wild, Andreas and Orchard, Garrick and Sandamirskaya, Yulia and Guerra, Gabriel A Fonseca and Joshi, Prasad and Plank, Philipp and Risbud, Sumedh R},
  journal={Proceedings of the IEEE},
  volume={109},
  number={5},
  pages={911--934},
  year={2021},
  publisher={IEEE}
}

@article{liu2021low,
  title={Low-power computing with neuromorphic engineering},
  author={Liu, Dingbang and Yu, Hao and Chai, Yang},
  journal={Advanced Intelligent Systems},
  volume={3},
  number={2},
  pages={2000150},
  year={2021},
  publisher={Wiley Online Library}
}

@article{modha2023neural,
  title={Neural inference at the frontier of energy, space, and time},
  author={Modha, Dharmendra S and Akopyan, Filipp and Andreopoulos, Alexander and Appuswamy, Rathinakumar and Arthur, John V and Cassidy, Andrew S and Datta, Pallab and DeBole, Michael V and Esser, Steven K and Otero, Carlos Ortega and others},
  journal={Science},
  volume={382},
  number={6668},
  pages={329--335},
  year={2023},
  publisher={American Association for the Advancement of Science}
}

@article{wu2022training,
  title={Training spiking neural networks for reinforcement learning tasks with temporal coding method},
  author={Wu, Guanlin and Liang, Dongchen and Luan, Shaotong and Wang, Ji},
  journal={Frontiers in Neuroscience},
  volume={16},
  pages={877701},
  year={2022},
  publisher={Frontiers Media SA}
}

@article{akl2023toward,
  title={Toward robust and scalable deep spiking reinforcement learning},
  author={Akl, Mahmoud and Ergene, Deniz and Walter, Florian and Knoll, Alois},
  journal={Frontiers in Neurorobotics},
  volume={16},
  pages={1075647},
  year={2023},
  publisher={Frontiers Media SA}
}

@article{xiao2022online,
  title={Online training through time for spiking neural networks},
  author={Xiao, Mingqing and Meng, Qingyan and Zhang, Zongpeng and He, Di and Lin, Zhouchen},
  journal={Advances in neural information processing systems},
  volume={35},
  pages={20717--20730},
  year={2022}
}

@article{zhu2022training,
  title={Training spiking neural networks with event-driven backpropagation},
  author={Zhu, Yaoyu and Yu, Zhaofei and Fang, Wei and Xie, Xiaodong and Huang, Tiejun and Masquelier, Timoth{\'e}e},
  journal={Advances in Neural Information Processing Systems},
  volume={35},
  pages={30528--30541},
  year={2022}
}

@article{yin2023accurate,
  title={Accurate online training of dynamical spiking neural networks through forward propagation through time},
  author={Yin, Bojian and Corradi, Federico and Boht{\'e}, Sander M},
  journal={Nature Machine Intelligence},
  volume={5},
  number={5},
  pages={518--527},
  year={2023},
  publisher={Nature Publishing Group UK London}
}

@book{amari2016information,
  title={Information geometry and its applications},
  author={Amari, Shun-ichi},
  volume={194},
  year={2016},
  publisher={Springer}
}

@article{nielsen2020elementary,
  title={An elementary introduction to information geometry},
  author={Nielsen, Frank},
  journal={Entropy},
  volume={22},
  number={10},
  pages={1100},
  year={2020},
  publisher={MDPI}
}

@article{SuperSpike,
  title={Superspike: Supervised learning in multilayer spiking neural networks},
  author={Zenke, Friedemann and Ganguli, Surya},
  journal={Neural computation},
  volume={30},
  number={6},
  pages={1514--1541},
  year={2018},
  publisher={MIT Press One Rogers Street, Cambridge, MA 02142-1209, USA journals-info~…}
}

@article{Esser2016ConvolutionalNF,
  title={Convolutional networks for fast, energy-efficient neuromorphic computing},
  author={Steven K. Esser and Paul Merolla and John V. Arthur and Andrew S. Cassidy and Rathinakumar Appuswamy and Alexander Andreopoulos and David J. Berg and Jeffrey L. McKinstry and Timothy Melano and Davis Barch and Carmelo di Nolfo and Pallab Datta and Arnon Amir and Brian Taba and Myron Flickner and Dharmendra S. Modha},
  journal={Proceedings of the National Academy of Sciences},
  year={2016},
  volume={113},
  pages={11441 - 11446}
}

@article{zenke2021remarkable,
  title={The remarkable robustness of surrogate gradient learning for instilling complex function in spiking neural networks},
  author={Zenke, Friedemann and Vogels, Tim P},
  journal={Neural computation},
  volume={33},
  number={4},
  pages={899--925},
  year={2021},
  publisher={MIT Press One Rogers Street, Cambridge, MA 02142-1209, USA journals-info~…}
}

@misc{snn_initialization,
title={Accelerating Training of Deep Spiking Neural Networks with Parameter Initialization},
author={Jianhao Ding and Jiyuan Zhang and Zhaofei Yu and Tiejun Huang},
year={2022},
url={https://openreview.net/forum?id=T8BnDXDTcFZ}
}
\bibliographystyle{icml2026}

\clearpage
\appendix

\renewcommand{\thefigure}{S\arabic{figure}}
\setcounter{figure}{0}
\renewcommand{\thetable}{S\arabic{table}}
\setcounter{table}{0}

\section{Comparison of different model architecture}

Table~\ref{table:architect} compares the model architectures used in our experiments in terms of parameter count, numerical precision, memory footprint, and dominant per-step operations.
We design these architectures to support a fair comparison along two axes: (i) \emph{algorithmic training rule} (SATR vs.\ EC/ES/SG/PPO) and (ii) \emph{system implementation} (bitset-accelerated vs.\ conventional dense execution).

\paragraph{RSNN variants.}
All RSNN-based methods share the same recurrent spiking policy backbone (a single recurrent spiking layer with linear read-in/read-out), resulting in an identical hidden size ($256$) and comparable parameter count ($\approx$193K).
The difference between \textbf{RSNN (EC)} and \textbf{RSNN* (SATR)} is \emph{not} the network structure, but the execution backend:
RSNN (EC) uses a conventional dense implementation whose dominant cost is matrix multiplication, whereas RSNN* (SATR) employs our bitset engine that packs binary spikes and binary connectivity/weights into machine words and computes synaptic integration via \texttt{AND}+\texttt{popcount} (Sec.~\ref{subsec:bitset_impl}).
This change preserves the underlying objective and policy class while significantly reducing memory traffic and wall-clock runtime.

\paragraph{Precision and memory footprint.}
For EC/SATR, the optimized variables are Bernoulli connectivity parameters and sampled connectivities are binary; combined with binary spikes, the core RSNN state can be represented in 1-bit precision.
Consequently, RSNN and RSNN* have a compact model footprint (24KB in Table~\ref{table:architect}), making them suitable for deployment on memory-constrained platforms and aligning with neuromorphic execution.
In contrast, the ES- and SG-trained RSNN baselines use FP32 weights in their standard formulations, increasing storage by $\sim$32$\times$ (768KB) and relying on floating-point matrix multiplications.

\paragraph{LSTM baseline parameter matching.}
To compare against a strong gradient-based recurrent policy, we include an LSTM trained with PPO.
We set the LSTM hidden size to $128$ so that its total parameters ($\approx$191K) closely match the RSNN parameter budget ($\approx$193K), controlling for representational capacity and ensuring that performance differences are not primarily driven by model size.
Despite comparable parameter counts, the LSTM requires FP32 storage (764KB) and dense matrix multiplications, which typically leads to higher runtime and energy consumption than binary RSNN execution.

Overall, Table~\ref{table:architect} highlights that SATR-RSNN achieves competitive control performance with a model that is simultaneously compact (1-bit), compute-efficient (bitwise operations), and comparable in parameter count to standard recurrent baselines, supporting the reward--runtime advantages observed in Fig.~\ref{fig:reward_runtime}.

\begin{table}[h!]
    \caption{Comparison of Model Architectures}
    \centering
    \resizebox{\columnwidth}{!}{
    \begin{tabular}{lccccc}
        \toprule
        \textbf{Model} & \textbf{Hidden size} & \textbf{Params} & \textbf{Precision}&  \textbf{Size (KB)}& \textbf{Operation} \\
        \midrule
        RSNN* (SATR) & 256 & 193K & 1-bit & 24 & $AND+Popcount$ \\
        RSNN (EC) & 256& 193K & 1-bit & 24 & $Matmul$ \\
        RSNN (ES\&SG) & 256 & 193K & FP32 & 768 & $Matmul$\\
        LSTM & 128 & 191K & FP32 & 764 & $Matmul$\\
        \bottomrule
    \end{tabular}}
    \label{table:architect}
\end{table}

\section{On-Chip Energy Consumption Estimation}
\label{app:power}

We estimate the \emph{on-chip} energy consumption of SATR by counting spiking-network operations and multiplying by published energy-per-operation values reported for Loihi~\citep{Loihi2}, which we use as a representative neuromorphic reference.
All parameters used in the estimate are summarized in Table~\ref{tab:energy}.

\begin{table}[h!]
    \centering
    \caption{Energy-per-operation and network/training parameters used for the analytical on-chip energy estimate.}
    \begin{tabular}{lc}
        \toprule
        \textbf{Parameter} & \textbf{Value}\\
        \midrule
        Energy per synaptic spike op $P_s$ & 23.6\,pJ \\
        Within-tile spike energy $P_w$ & 1.7\,pJ \\
        Energy per neuron update $P_u$ & 81\,pJ \\
        \midrule
        \# Generations $G$ & 2000 \\
        Population size $P$ & $\{2^{10},\dots,2^{13}\}$ \\
        \# Time steps per rollout $S$ & 33{,}200 \\
        \# Neurons $N$ & 256 \\
        Avg. spikes per neuron per step $R$ & 0.025 \\
        \# Connections per neuron $C$ & 128 \\
        \# Update ops per neuron per step $I$ & 4 \\
        \bottomrule
    \end{tabular}
    \label{tab:energy}
\end{table}

\paragraph{Energy per policy evaluation (one rollout).}
We define $E_{\text{one}}$ as the energy to execute one RSNN policy evaluation over $S$ discrete timesteps.
At each timestep, we account for: (i) neuron state updates and (ii) synaptic/spike-processing events.
With $N$ neurons, $I$ update operations per neuron per timestep, and energy $P_u$ per neuron update, the neuron-update energy is:
\[
E_{\text{upd}} = P_u \, N \, I \, S.
\]
We model spiking activity via an average spike rate $R$ (spikes per neuron per timestep), yielding an expected $NRS$ spikes per rollout.
For each spike, we include the synaptic processing energy $P_s$ plus an intra-tile spike delivery energy $P_w$ amortized over $C$ connections per neuron, giving:
\[
E_{\text{syn}} = \left(P_s + C P_w\right) N R S.
\]
The total energy per rollout is therefore
\begin{equation}
E_{\text{one}} = E_{\text{upd}} + E_{\text{syn}}
= P_u N I S + \left(P_s + C P_w\right) N R S
\approx 2.8\,\text{mJ}.
\label{eq:Eone}
\end{equation}

\paragraph{Total energy over training.}
SATR evaluates a population of $P$ individuals over $G$ generations.
Assuming each individual requires one rollout evaluation, the total on-chip energy is
\begin{equation}
E_{\text{tot}} = E_{\text{one}} \, G \, P,
\label{eq:Etot}
\end{equation}
which consequently yields \textbf{5.7--45.6\,kJ} for $P \in \{1024, 2048, 4096, 8192\}$ with the Humanoid settings in Table~\ref{tab:power_consumption}.

\begin{table}[h!]
    \caption{Estimated total energy consumption for SATR-based RSNN training on Humanoid under different population sizes, compared with a PPO-LSTM baseline. }
    \centering
    \resizebox{\columnwidth}{!}{
    \begin{tabular}{lccccc}
        \toprule
        \textbf{Population size $P$} & 1024 & 2048 & 4096 & 8192 & PPO-LSTM \\
        \midrule
        Estimated RSNN on-chip energy [kJ] & 5.7 & 11.4 & 22.8 & 45.6 & -- \\
        Measured GPU energy [MJ] & -- & -- & -- & -- & 18.4 \\
        \bottomrule
    \end{tabular}}
    \label{tab:power_consumption}
\end{table}

\paragraph{GPU baseline.}
For reference, PPO-LSTM training on the same task consumed \textbf{18.4\,MJ} of measured GPU energy.
The gap is consistent with the sparse, event-driven computation of RSNNs versus dense floating-point operations in conventional deep RL.

\paragraph{Limitations.}
These calculations are analytical and intended for order-of-magnitude comparison.
They rely on published per-operation energy figures~\citep{Loihi2} and on the operation counts implied by Table~\ref{tab:energy}.
Future work will validate these estimates via direct experiments on neuromorphic hardware.

\section{Performance Comparison with PPO and SG}
\label{app:ppo}

\subsection{Full PPO Results}
To further evaluate the effectiveness of the EC-RSNN approach in comparison with contemporary deep reinforcement learning (RL) methods, we conducted additional training of LSTM models using Proximal Policy Optimization (PPO) \citep{schulman2017proximal} on the Humanoid task, which constitutes the most challenging environment in our benchmark suit. To ensure a fair comparison under comparable computational budgets, the training runtime of PPO-LSTM was matched to different population settings of $SATR$-$RSNN^\ast$. Specifically, three PPO-LSTM variants were trained with 500k iterations (corresponding to the runtime of SATR-pop4096), 850k iterations (corresponding to the runtime of SATR-pop8192), and 2.2M iterations (corresponding to the runtime of SATR-pop8192 without bitset acceleration).

\textbf{As an additional sanity check on PPO training sufficiency}, we compare our final returns to the tuned PPO performance reported by a \textbf{Brax maintainer on Humanoid} (11,300 average return over 3 seeds) in \url{https://github.com/google/brax/discussions/230#discussioncomment-3879848}. Our 500k-iteration run already reaches 11,326.8, and longer training further improves performance to 12,396.8 (850k) and 12,876.0 (2.2M), indicating our PPO baseline is not under-trained.

\begin{figure}[ht]
    \centering
    \includegraphics[width=\linewidth]{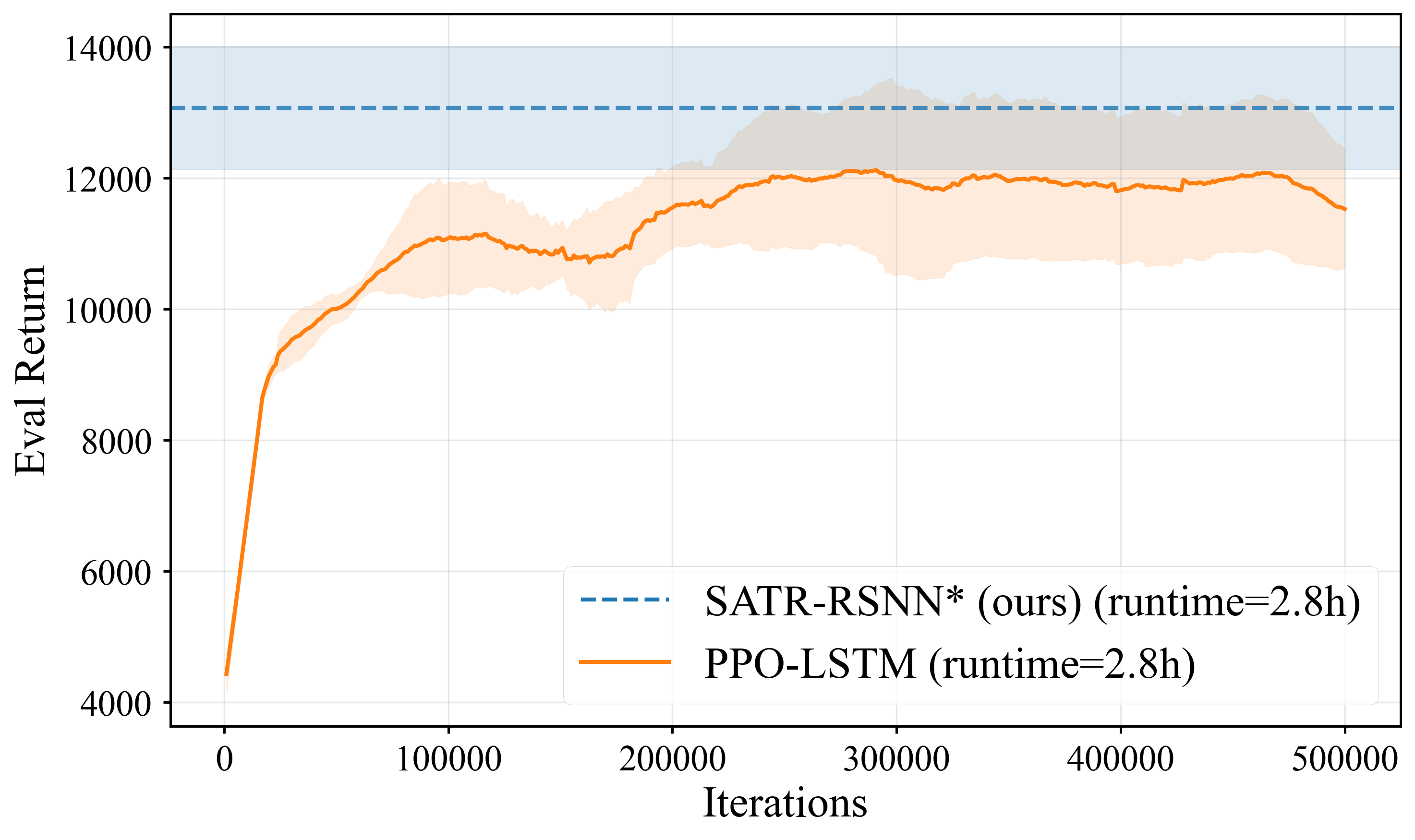}
        \caption{Performance comparison between PPO-LSTM and SATR-RSNN* (population = 4096) on the Humanoid task under matched computational budgets. Learning curves are smoothed for visualization.}
    \label{fig:ppo_500k}
\end{figure}
\begin{figure}[ht]
    \centering
    \includegraphics[width=\linewidth]{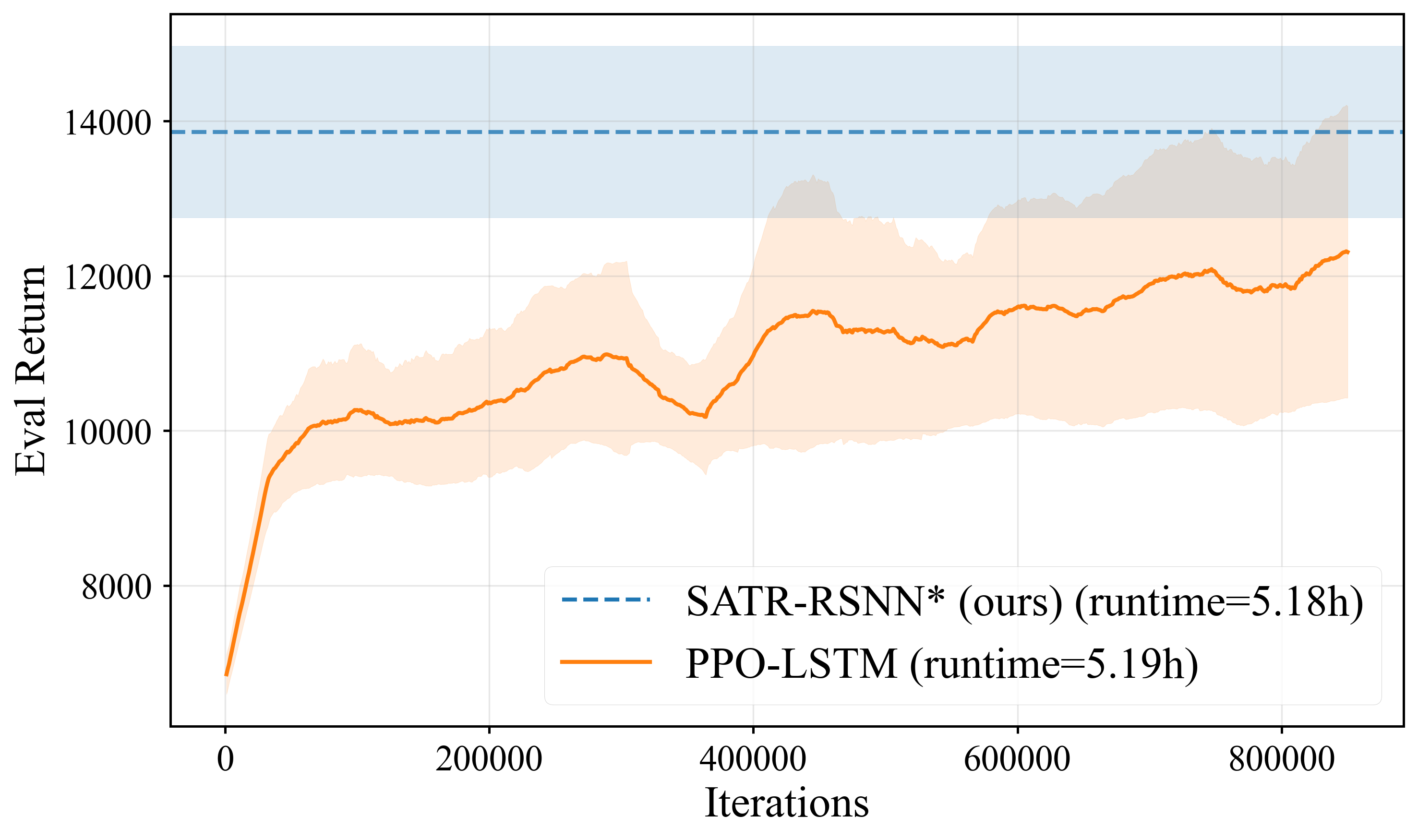}
    \caption{Performance comparison between PPO-LSTM and SATR-RSNN* (population = 8192) on the Humanoid task under matched computational budgets. Learning curves are smoothed for visualization.}
    \label{fig:ppo_850k}
\end{figure}
\begin{figure}[ht]
    \centering
    \includegraphics[width=\linewidth]{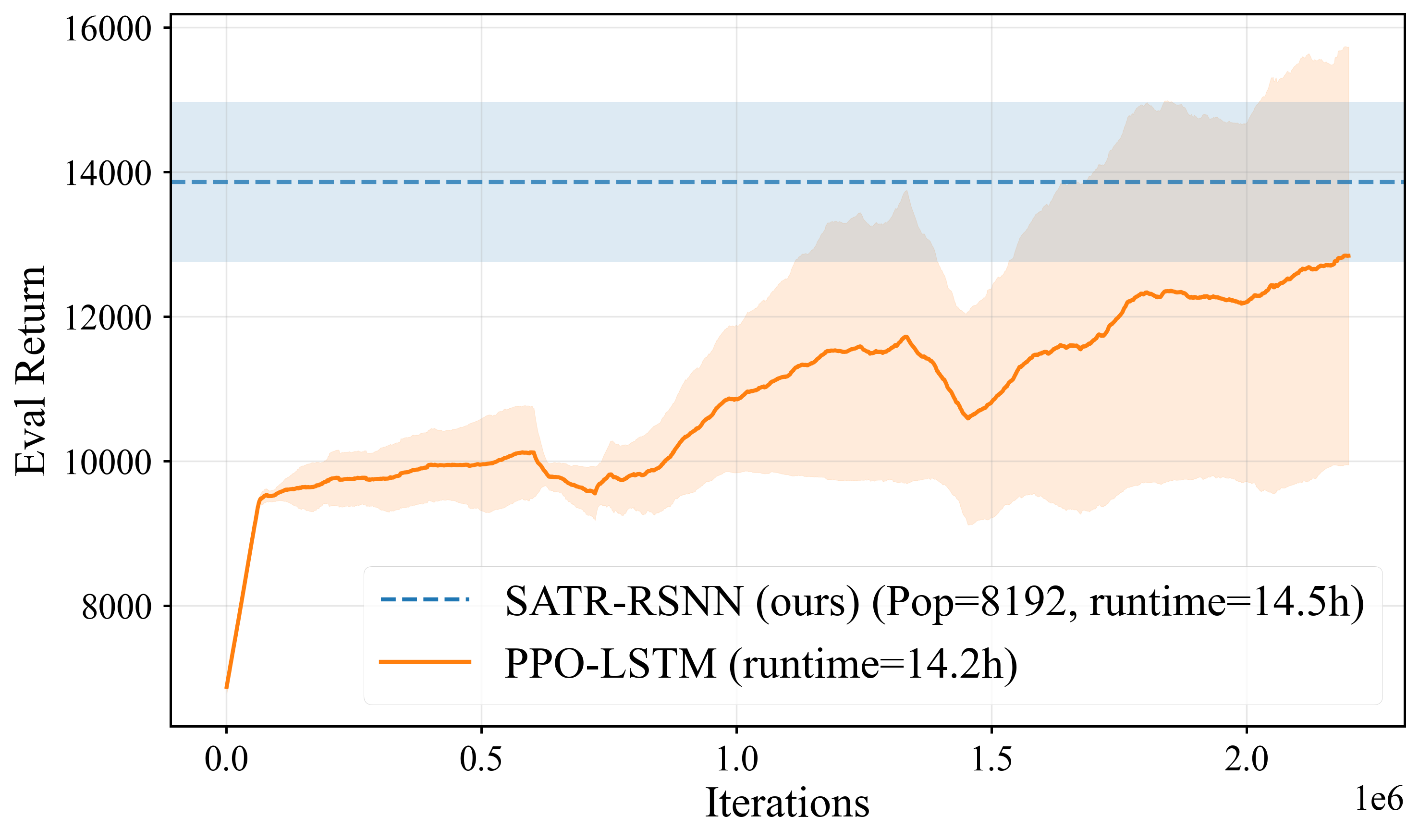}
    \caption{Performance comparison between PPO-LSTM and SATR-RSNN* \textbf{without acceleration} (population = 8192) on the Humanoid task under matched computational budgets. Learning curves are smoothed for visualization.}
    \label{fig:ppo_2200k}
\end{figure}

The resulting learning curves are shown in Figures~\ref{fig:ppo_500k}, ~\ref{fig:ppo_850k} and \ref{fig:ppo_2200k}. Under these runtime-matched conditions, SATR-RSNN* achieves higher returns and demonstrates more stable learning dynamics than PPO-LSTM on the Humanoid task. These results indicate that SATR-RSNN* attains performance comparable to strong gradient-based deep RL baselines, while relying on binary spiking dynamics and gradient-free optimization.


\subsection{Full SG Results}
\label{app:full_sg_results}

We present a comprehensive analysis of the Surrogate Gradient (SG) results, considering various damping factors ($\gamma$) and the parameter ($\beta$). The outcomes are illustrated in Figure \ref{fig:exp_CurveOfSG}.

\renewcommand{\thefigure}{S\arabic{figure}}

\subsection{Training Objective vs.\ Evaluation Metric}
\label{app:train-vs-eval}

We distinguish the \emph{training objective} optimized by each algorithm from the
\emph{evaluation metric} reported throughout the paper.
Some baselines (e.g., PPO) optimize a discounted surrogate objective (e.g., with $\gamma<1$ and GAE),
whereas our method optimizes an undiscounted objective ($\gamma=1$).
This difference is intrinsic to the algorithms and does not affect the comparability of the reported
performance numbers, because \textbf{all methods are evaluated using the same metric under an identical
evaluation protocol}:
\begin{figure*}[ht]
  \centering
  \includegraphics[width=\textwidth]{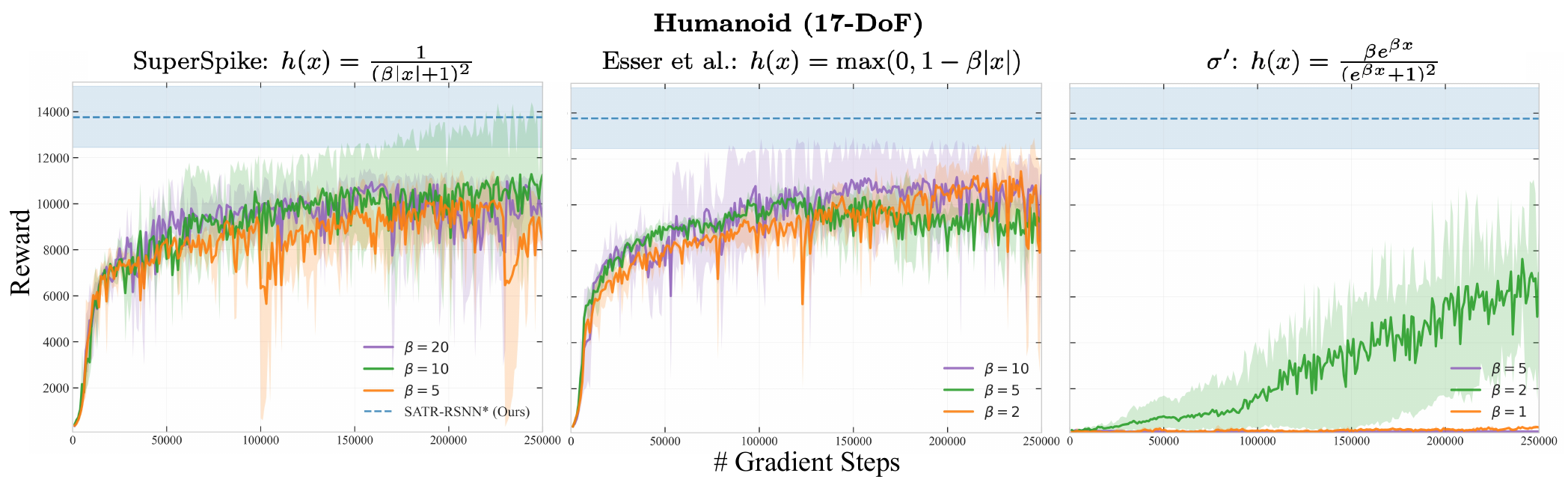}
  \caption{Performance of different surrogate gradient methods with different training parameters in the Humanoid task, which indicates our method beats the best parameter combination of surrogate gradient methods.}
  \label{fig:exp_CurveOfSG}
\end{figure*}
\begin{itemize}
    \item \textbf{Evaluation policy.} At evaluation time, we take the \emph{current} policy of each method
    (no parameter averaging unless explicitly stated).
    \item \textbf{Evaluation rollouts.} We run \textbf{128 parallel environments} with this policy in inference
    mode (no learning updates during evaluation).
    \item \textbf{Metric.} For each rollout, we compute the \textbf{undiscounted episodic return} from the raw
    environment rewards and report the \textbf{mean return across the 128 episodes}.
    \item \textbf{Episode termination.} We follow the environment's standard termination conditions and horizon.
\end{itemize}

Note that $\gamma$ only affects training-time credit assignment and the surrogate objective;
it does not enter the evaluation metric, which is computed from the raw environment rewards.
Therefore, while training objectives may differ, the reported numbers measure the same quantity:
the expected episodic performance of the learned policy under a standardized evaluation setting.
\section{Derivation of the EC+TR Closed-Form Step}
\label{app:ec_tr_derivation}

This appendix derives the analytic update used by the fixed-budget trust-region ablation added to the EC baseline (EC+TR).
Our trust region is defined over the search distribution $p_\rho(\theta)$, not over an empirical policy distribution.
Under the local quadratic (Fisher) approximation of the KL divergence, the resulting trust-region subproblem admits a
closed-form solution, eliminating the need for TRPO-style line search/bisection.

\subsection{KL trust region in distribution space}

EC+TR constrains the KL change between the current and updated search distributions:
\begin{equation}
D_{\mathrm{KL}}\!\big(p_{\rho}\,\|\,p_{\rho'}\big) \le 2\delta,
\qquad \rho' = \rho + \Delta\rho .
\label{eq:app_ec_tr_constraint}
\end{equation}
Using the standard second-order expansion of KL around $\rho$,
\begin{equation}
D_{\mathrm{KL}}\!\big(p_{\rho}\,\|\,p_{\rho+\Delta\rho}\big)
=
\frac{1}{2}\,\Delta\rho^\top F(\rho)\,\Delta\rho
\;+\; o(\|\Delta\rho\|^2),
\label{eq:app_ec_tr_kl_quad}
\end{equation}
where $F(\rho)$ is the Fisher information matrix of the distribution family $\{p_\rho\}$,
\begin{equation}
F(\rho)
=
\mathbb{E}_{\theta\sim p_\rho}\!\left[
\nabla_\rho \log p_\rho(\theta)\,\nabla_\rho \log p_\rho(\theta)^\top
\right],
\label{eq:app_ec_tr_fisher}
\end{equation}
the constraint \eqref{eq:app_ec_tr_constraint} reduces (up to second order) to the quadratic trust region
\begin{equation}
\frac{1}{2}\,\Delta\rho^\top F(\rho)\,\Delta\rho \le \delta.
\label{eq:app_ec_tr_quadratic_constraint}
\end{equation}

\subsection{Trust-region subproblem with a linearized objective}

Let $\mathcal{L}(\rho)$ denote the objective optimized by EC in distribution-parameter space (e.g., an expected fitness or
a surrogate improvement). We linearize $\mathcal{L}$ locally:
\begin{equation}
\mathcal{L}(\rho+\Delta\rho)
\approx
\mathcal{L}(\rho) + \tilde{g}^\top \Delta\rho,
\label{eq:app_ec_tr_linear_obj}
\end{equation}
where $\tilde{g}$ is the local ascent direction in the Euclidean parameterization.
Then EC+TR solves
\begin{equation}
\max_{\Delta\rho}\quad \tilde{g}^\top \Delta\rho
\quad \text{s.t.}\quad
\frac{1}{2}\,\Delta\rho^\top F(\rho)\,\Delta\rho \le \delta.
\label{eq:app_ec_tr_subproblem}
\end{equation}
This is a standard trust-region problem in the local KL geometry induced by $F(\rho)$.

\subsection{Closed-form normalized natural step}

Form the Lagrangian
\begin{equation}
\mathcal{J}(\Delta\rho,\lambda)
=
\tilde{g}^\top \Delta\rho
-\lambda\left(\frac{1}{2}\Delta\rho^\top F\Delta\rho - \delta\right),
\qquad \lambda \ge 0.
\label{eq:app_ec_tr_lagrangian}
\end{equation}
Stationarity w.r.t.\ $\Delta\rho$ yields
\begin{equation}
\nabla_{\Delta\rho}\mathcal{J}
=
\tilde{g} - \lambda F\Delta\rho
=
0
\quad\Rightarrow\quad
\Delta\rho = \frac{1}{\lambda}\,F^{-1}\tilde{g}.
\label{eq:app_ec_tr_stationary}
\end{equation}
At optimum the constraint is active (unless $\tilde{g}=0$), hence
\begin{equation}
\frac{1}{2}\Delta\rho^\top F\Delta\rho = \delta.
\label{eq:app_ec_tr_active}
\end{equation}
Substituting \eqref{eq:app_ec_tr_stationary} into \eqref{eq:app_ec_tr_active} gives
\begin{equation}
\frac{1}{2}\left(\frac{1}{\lambda}F^{-1}\tilde{g}\right)^\top
F
\left(\frac{1}{\lambda}F^{-1}\tilde{g}\right)
=
\frac{1}{2\lambda^2}\,\tilde{g}^\top F^{-1}\tilde{g}
=
\delta,
\end{equation}
so that
\begin{equation}
\lambda
=
\sqrt{\frac{\tilde{g}^\top F^{-1}\tilde{g}}{2\delta}}.
\label{eq:app_ec_tr_lambda}
\end{equation}
Plugging back yields the closed-form normalized natural-gradient step:
\begin{equation}
\Delta\rho
=
\sqrt{2\delta}\;
\frac{F^{-1}\tilde{g}}{\sqrt{\tilde{g}^\top F^{-1}\tilde{g}}}.
\label{eq:app_ec_tr_closed_form_standard}
\end{equation}

\paragraph{Consistency with the main-text notation.}
In our implementation and main text, we denote by
\begin{equation}
g \;\triangleq\; F^{-1}\tilde{g}
\label{eq:app_ec_tr_precond_def}
\end{equation}
the Fisher-preconditioned (natural) ascent direction in distribution space.
Under this convention, \eqref{eq:app_ec_tr_closed_form_standard} becomes
\begin{equation}
\Delta\rho
=
\sqrt{2\delta}\;
\frac{g}{\sqrt{g^\top F^{-1} g}},
\label{eq:app_ec_tr_closed_form_maintext}
\end{equation}
which is exactly the update used by EC+TR in the main text.
Moreover, \eqref{eq:app_ec_tr_closed_form_maintext} satisfies the quadratic trust-region constraint
$\tfrac{1}{2}\Delta\rho^\top F\Delta\rho = \delta$ (up to the local KL approximation).

\subsection{Bernoulli Fisher for factorized distributions}

For the factorized Bernoulli family
$p_\rho(\theta)=\prod_{i=1}^d \mathrm{Bernoulli}(\theta_i;\rho_i)$ with $\rho_i\in(0,1)$,

\begin{equation}
\log p_\rho(\theta)
=
\sum_{i=1}^d\left[\theta_i\log\rho_i+(1-\theta_i)\log(1-\rho_i)\right],
\end{equation}
and
\begin{equation}
\frac{\partial}{\partial\rho_i}\log p_\rho(\theta)
=
\frac{\theta_i-\rho_i}{\rho_i(1-\rho_i)}.
\end{equation}
Therefore the Fisher information is diagonal:
\begin{equation}
F_{ij}(\rho) = 
\begin{cases} 
\frac{1}{\rho_i(1-\rho_i)}, & i = j \\
0, & i \neq j
\end{cases}
\label{eq:app_ec_tr_bernoulli_fisher}
\end{equation}

Substituting \eqref{eq:app_ec_tr_bernoulli_fisher} into \eqref{eq:app_ec_tr_closed_form_maintext}
yields an element-wise scaled step for EC+TR.

\subsection{Why EC+TR does not require line search}

Canonical TRPO often relies on backtracking/line search because the KL constraint is enforced
on an empirical policy distribution (over visited states) and may not be satisfied by approximate updates
under sampling noise and function approximation.
In contrast, EC+TR defines the KL directly over $p_\rho(\theta)$ and uses the local quadratic KL approximation
\eqref{eq:app_ec_tr_kl_quad}, which turns the constraint into the quadratic form
\eqref{eq:app_ec_tr_quadratic_constraint}. The resulting trust-region subproblem
\eqref{eq:app_ec_tr_subproblem} has the analytic solution \eqref{eq:app_ec_tr_closed_form_maintext},
so the step size is obtained in closed form and no bisection/line search is needed.

\section{Hardware}

All experiments were conducted on a single GPU server with the following specifications. Additionally, all efficiency experiments, including those that measure wall-clock time, were executed on a single GPU within this server.

\begin{itemize}
\item 1$\times$ NVIDIA GeForce RTX 4090
\item 2$\times$ Intel Xeon Silver 4110 CPUs
\item 252GB system  Memory
\end{itemize}

\section{Hyperparameters}

In this section, we provide the hyperparameters for our framework, Evolving Connectivity (EC), and the baselines, including Evolution Strategies (ES) and Surrogate Gradient (SG). All methods utilize the same hyperparameter set for all three locomotion tasks. We also list the settings of neural networks below.

\begin{table}[H]
    \caption{Hyperparameters for SATR (ours) and Evolving Connectivity (EC)}
    \centering
    \begin{tabular}{lc}
        \toprule
        \textbf{Hyperparameter} & \textbf{Value} \\
        \midrule
        Learning rate $\eta$ & 0.15 \\
        Exploration probability $\epsilon$ & $10^{-3}$ \\
        \bottomrule
    \end{tabular}
\end{table}

\begin{table}[H]
    \caption{Hyperparameters for Evolution Strategies}
    \centering
    \begin{tabular}{lcc}
        \toprule
        \textbf{Hyperparameter} & \textbf{Value}  \\
        \midrule
        Learning rate $\eta$ & 0.15 \\

        Noise standard deviation $\sigma$ & 0.3 \\
        Weight decay & 0.1  \\
        \bottomrule
    \end{tabular}
\end{table}

\begin{table}[H]
    \caption{Hyperparameters for Surrogate Gradient}
    \centering
    \begin{tabular}{lc}
        \toprule
        \textbf{Hyperparameter} & \textbf{Value} \\
        \midrule
        \multicolumn{2}{c}{\textit{Proximal Policy Optimization (PPO)}} \\
        \midrule
        Batch size & 2048 \\
        BPTT length & 16 \\
        Learning rate $\eta$ & $3 \times 10^{-4}$ \\
        Clip gradient norm & 0.5 \\
        Discount $\gamma$ for \textbf{Training} & 0.99 \\
        Discount $\gamma$ for \textbf{Evaluation} & 1 \\
        GAE $\lambda$ & 0.95 \\
        PPO clip & 0.2 \\
        Value loss coefficient & 1.0 \\
        Entropy coefficient & $10^{-3}$ \\
        \midrule
        \multicolumn{2}{c}{\textit{Surrogate Gradient}} \\
        \midrule
        Surrogate function & AbsBeta \\
        Surrogate function parameter $\beta$ & 10 \\
        \bottomrule
    \end{tabular}
\end{table}

\begin{table}[H]
    \caption{Settings of neural networks}
    \centering
    \begin{tabular}{lc}
        \toprule
        \textbf{Hyperparameter} & \textbf{Value} \\
        \midrule
        \multicolumn{2}{c}{\textit{Recurrent Spiking Neural Network (RSNN)}} \\
        \midrule
        Number of neurons $d_h$ & 256 \\
        Excitatory ratio & $50\%$ \\
        Simulation time per environment step & 16.6 ms \\
        Simulation timestep $\Delta t$ & 0.5 ms \\
        Synaptic time constant $\tau_{syn}$ & 5.0 ms \\
        Membrane time constant $\tau_{m}$ & 10.0 ms \\
        Output time constant $\tau_{out}$ & 10.0 ms \\
        Input membrane resistance \footnotemark[1] $R_{in}$ & $0.15 \cdot \tau_m \sqrt{\frac{2}{d_{in}}}$ \\
        Hidden membrane resistance \footnotemark[1] $R_{h}$ & $1.0 \cdot \frac{\tau_m}{\tau_{syn}} \sqrt{\frac{2}{d_{h}}}$ \\
        Output membrane resistance \footnotemark[1] $R_{out}$ & $5.0 \cdot \tau_{out} \sqrt{\frac{2}{d_{h}}}$ \\
        \midrule
        \multicolumn{2}{c}{\textit{Long-short term memory (LSTM)}} \\
        \midrule
        Hidden size & 128 \\
        \bottomrule
    \end{tabular}
\end{table}

\section{Discussion}
\label{app:discussion}

SATR targets a practical failure mode of population-based RSNN training: with finite populations, noisy updates can cause overly large \emph{distributional} jumps, especially for sparse Bernoulli connectivity where KL curvature explodes near $\rho\!\to\!0/1$.
By tying the allowable KL displacement to the signal energy $\|g\|_2^2$, SATR expands when the update direction is coherent and contracts when it is noise-dominated, which explains its stability advantage in small-population regimes (Table~\ref{tab:all_performance}) and its reduced sensitivity relative to fixed-budget trust regions (Table~\ref{tab:ablation_tr}).
Specializing to Bernoulli yields the boundary-aware scaling $\sqrt{\rho(1-\rho)}$ (Eq.~\eqref{eq:satr_ec_update}), preventing curvature-induced KL blow-ups that can destabilize EC.

SATR improves optimization robustness, while our bitset backend improves throughput without changing the learning rule.
Because connectivity and spikes are binary, \texttt{AND}+\texttt{popcount} replaces dense matmuls, yielding substantial wall-clock gains and a better reward--runtime trade-off (Fig.~\ref{fig:reward_runtime}, Table~\ref{tab:ablation_acceleration}).

\section{Limitation and future work.}
Limitations include reliance on the local (Fisher) KL approximation, the factorized distribution assumption, and energy results being analytical rather than measured on neuromorphic hardware.
Future work includes extending SATR to structured connectivity distributions, incorporating uncertainty estimates of $g$ to better calibrate trust-region size, and validating energy/latency on neuromorphic deployments.

\footnotetext[1]{Resistance is set following~\cite{snn_initialization} to preserve variance.}

\end{document}